\documentclass[journal]{IEEEtran}
\usepackage{times}
\usepackage[ruled]{algorithm2e} 
\usepackage{epsfig}
\usepackage{graphicx}
\usepackage{amsmath}
\usepackage{amssymb}
\usepackage{booktabs}
\usepackage{multirow}
\usepackage{enumitem}
\usepackage[pagebackref=false,breaklinks=true,letterpaper=true,colorlinks,urlcolor=blue,citecolor=blue,linkcolor=blue,bookmarks=false]{hyperref}

\usepackage{cite}

\ifCLASSINFOpdf
\else
\fi
\begin{document}
%
\title{Real-Time Correlation Tracking via Joint Model Compression and Transfer}
%
%

\author{Ning Wang, 
	    Wengang Zhou, 
	    Yibing Song,
	    Chao Ma
	    and~Houqiang Li
\thanks{Ning Wang, Wengang Zhou, and Houqiang Li are with the CAS Key Laboratory of Technology in Geo-spatial Information Processing and Application System, Department of Electronic Engineering and Information Science, University of Science and Technology of China, Hefei, China. \protect\\ E-mail: wn6149@mail.ustc.edu.cn, \{zhwg, lihq\}@ustc.edu.cn.}
\thanks{Yibing Song is with the Tencent AI Lab, Shenzhen, China.\protect\\ E-mail: dynamicstevenson@gmail.com.}
\thanks{Ma Chao is with the MoE Key Lab of Artificial Intelligence, AI Institute, Shanghai Jiao Tong University, Shanghai, China. \protect\\ E-mail: chaoma@sjtu.edu.cn.}
\thanks{Corresponding authors: Wengang Zhou and Houqiang Li.}}

%
%

\markboth{Journal of XXX,~Vol.~*, No.~*, July~2019}%
{Shell \MakeLowercase{\textit{et al.}}: Bare Demo of IEEEtran.cls for IEEE Journals}
%



\maketitle

\begin{abstract}
Correlation filters (CF) have received considerable attention in visual tracking because of their computational efficiency.
%
Leveraging deep features via off-the-shelf CNN models (e.g., VGG), CF trackers achieve state-of-the-art performance while consuming a large number of computing resources.
%
This limits deep CF trackers to be deployed to many mobile platforms on which only a single-core CPU is available.
In this paper, we propose to jointly compress and transfer off-the-shelf CNN models within a knowledge distillation framework.
%
We formulate a CNN model pretrained from the image classification task as a teacher network, and distill this teacher network into a lightweight student network as the feature extractor to speed up CF trackers.
In the distillation process, we propose a fidelity loss to enable the student network to maintain the representation capability of the teacher network.
Meanwhile, we design a tracking loss to adapt the objective of the student network from object recognition to visual tracking.
%
The distillation process is performed offline on multiple layers and adaptively updates the student network using a background-aware online learning scheme.
%
%
Extensive experiments on five challenging datasets demonstrate that the lightweight student network accelerates the speed of state-of-the-art deep CF trackers to real-time on a single-core CPU while maintaining almost the same tracking accuracy.
\end{abstract}

\begin{IEEEkeywords}
Long-term tracking, tracking-by-detection, re-detection, feature combination.
\end{IEEEkeywords}

\IEEEpeerreviewmaketitle

\vspace{+0.00in}

\section{Introduction} \label{sec:intro}

\IEEEPARstart{T}{here} has been an increasing demand for visual object tracking algorithms in numerous vision applications. Typical examples include video surveillance, human-computer interaction, and autonomous driving.
As a key component, tracking target objects in real-time plays a critical role in improving the overall efficiency of vision applications.
%
The visual tracking framework based on Correlation Filters (CF) has been widely investigated in \cite{MOSSE,KCF,zhang2013STC} because of the efficient correlation computation in the Fourier domain.
%
When integrated with CNN features, CF trackers \cite{HCF,ECO,UPDT} achieve state-of-the-art tracking accuracy.
However, extracting high-dimensional deep features brings in a huge computational cost and limits CF trackers to achieve real-time speed.
Although deep operations can be always accelerated by GPUs, deep CF trackers cannot be deployed on CPU-only devices, e.g., most intelligent mobile phones do not have GPUs. Let alone the huge power consumption and memory storage required by existing pretrained CNN models (e.g., VGG \cite{VGG}).
%
The challenges of using off-the-shelf CNN models (e.g., VGG \cite{VGG}) include huge demand for memory storage, heavy computational burden, and high power consumption.
%
%
It is therefore non-trivial to investigate how to accelerate deep CF trackers on a CPU platform to achieve real-time speed without suffering a significant drop in tracking accuracy.


\begin{figure}
	\centering
	\includegraphics[width=8.9cm]{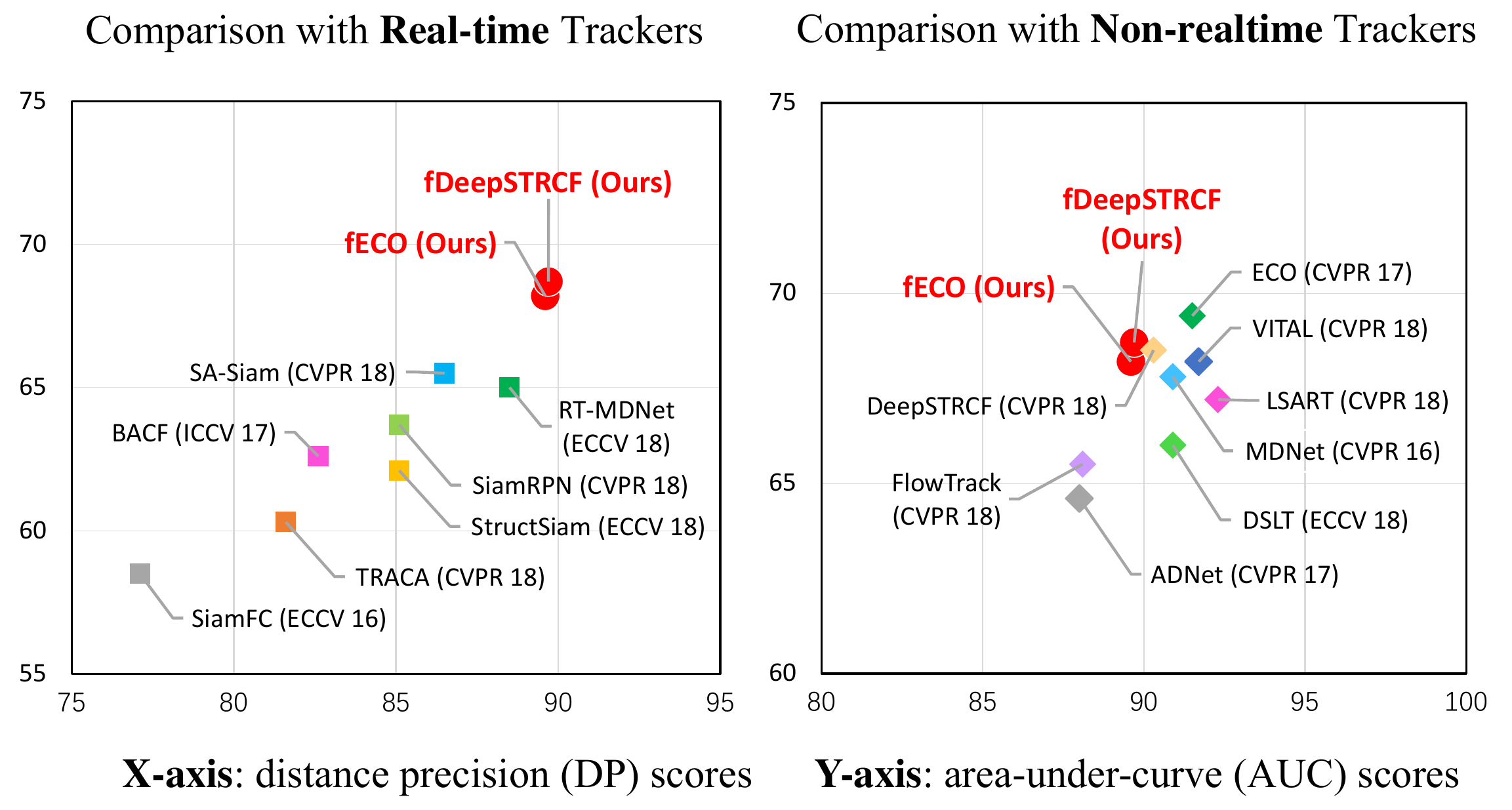}
	\caption{Tracking results on the OTB-2015 dataset \cite{OTB-2015}.
		%
		%
		%
		The proposed method accelerates state-of-the-art deep CF trackers (i.e., ECO \cite{ECO} and DeepSTRCF \cite{STRCF}) through joint CNN model compression and transfer.
		The improved CF trackers (i.e., fECO and fDeepSTRCF) perform favorably against existing methods and achieves real-time speed (more than 20 FPS) on a single-core CPU.
		It is worth mentioning that most existing real-time trackers as shown on the left cannot achieve real-time speed on a CPU, while the recent performance leaders shown on the right are far from real-time even on a GPU.}
	%
	\label{fig:1} \vspace{-0.0in}
\end{figure}


%
In this paper, we jointly compress and transfer off-the-shelf CNN models into a lightweight feature extractor.
The lightweight feature extractor enables deep CF trackers to achieve real-time speed as well as consume less memory.
The model compression and transfer are from the perspective of knowledge distillation \cite{hinton2015distilling,romero2014fitnets}.
We take the off-the-shelf model as a teacher network, which is pretrained for the object recognition task.
On the other side, a low capacity student network is used to learn from the teacher network.
%
%
In the distillation process, we propose two types of losses.
The first one is a fidelity loss and the second one is a correlation tracking loss.
The fidelity loss ensures the student network to convey the representation from the teacher network, while the correlation tracking loss transfers the objective of the student network from object recognition to visual tracking.
%
%
we take the hierarchies of deep models into account and perform the distillation process on multiple CNN layers offline.
After distillation, the student network maintains the high-level semantic discrimination from the fidelity loss.
Besides, the tracking loss helps the student network to produce target-specific CNN representations.
%
%
During online tracking, we propose a background-aware adaptation method to update the student network for further performance improvement.
%
%

The student network is a lightweight feature extraction backbone.
%
The model size of the student network is only 1.5 MB while the original size of the teacher network is 95 MB (i.e., $63\times$ smaller).
When integrated with the proposed lightweight backbone, the state-of-the-art deep CF trackers including ECO \cite{ECO} and DeepSTRCF \cite{STRCF} are able to achieve real-time speed on a single-core CPU while maintaining almost the same tracking accuracy on prevalent tracking benchmarks.

We summarize the contributions of our work as follows:
\begin{itemize}
	\setlength{\parskip}{0pt}	
	\item We compress and adapt off-the-shelf deep CNN models into lightweight backbones by knowledge distillation. We propose a fidelity loss and a correlation tracking loss to jointly compress the network and transfer its objective from object recognition to visual tracking.  	
	\item We propose to distillate student network via hierarchical CNN representations offline.
	We propose a background-aware adaption method to incrementally fine-tune the student network to adapt to target appearance changes.	
	\item We integrate the proposed lightweight backbone into the state-of-the-art deep CF trackers \cite{ECO,STRCF}. Evaluations on the large-scale benchmark datasets indicate the effectiveness of the proposed method in terms of the real-time speed and tracking accuracy.  
\end{itemize}

In the following of the paper, we describe the related work in Section \ref{sec:related}, correlation tracking in Section \ref{revisit}, the proposed approach in Section \ref{sec:proposed approach}, and experiments in Section \ref{sec:exp}. Finally, we conclude the paper in Section \ref{sec:conclusion}.

\section{Related Work}\label{sec:related}


In this section, we briefly survey the closely related literature on three aspects: tracking by correlation filters, real-time tracking, and network compression.

\subsection{Correlation Tracking}
Correlation filters have been widely studied in visual tracking since the MOSSE method \cite{MOSSE} was proposed by Bolme \emph{et al.} in 2010.
The correlation filter is trained by minimizing a ridge regression loss for all circular shifts of the training sample, which can be efficiently solved in the Fourier domain \cite{liu2018survey_tip}.
Heriques \emph{et al.} exploited the circulant structure of training patches in the kernel space \cite{KCF}. The SRDCF tracker \cite{SRDCF} alleviates the boundary effects by penalizing correlation filter coefficients depending on spatial locations. The CSR-DCF algorithm \cite{CSR-DCF} constructs filters with channel and spatial reliability. The C-COT \cite{C-COT} adopts a continuous-domain formulation and is further improved by an efficient convolution operator (ECO \cite{ECO}). The recent DRT tracker \cite{DRT} jointly learns the discrimination and reliability of CF. In addition, multiple kernels \cite{Tang_2018_multiKernel}, combination with particle filter \cite{zhang2017correlation_particle}, re-detection mechanism for long-term scenario \cite{LCT,ningwangTCSVT} and ensemble learning schemes \cite{Staple,MCCT,zhang2018parallel_tip} have also been investigated in the CF family. 
In recent years, the combination of CF trackers and deep features from off-the-shelf CNN models has demonstrated impressive results \cite{HCF,HDT,ECO}. Even though state-of-the-art results can be obtained by leveraging deep feature representations, the characteristic real-time efficiency of the correlation filter has gradually faded due to the adopted heavyweight CNN model. 
In this work, different from the above approaches putting emphasis on learning more discriminative filters, we focus on learning a distilled lightweight backbone network that enables high-performance real-time correlation tracking even on a single-core CPU.

\subsection{Real-time Tracking}
The Siamese network has been widely studied for real-time tracking.
The fully convolutional Siamese Network (SiamFC) regards the tracking task as a similarity learning problem, and compares the template patch with the candidate patches in the search patch in a sliding-window manner. 
On the basis of the SiamFC framework \cite{SiamFc}, the correlation layer \cite{CFNet}, attention mechanism \cite{RASNet}, semantic branch \cite{SASiam} and unsupervised learning scheme \cite{UDT} are widely explored. The recent region proposal Siamese network \cite{SiamRPN,DaSiamRPN} achieves higher speed compared with SiamFC \cite{SiamFc} by discarding multiple-scale estimation. However, the Siamese networks heavily rely on powerful GPUs and the running speed on CPU is only 2$ \sim $3 FPS \cite{EAST} due to heavyweight model complexity.

On the other hand, CF trackers can achieve real-time speed when using lightweight hand-crafted features such as HOG and ColorNames \cite{KCF,DSST,STRCF,MCCT}, but they typically have an obvious performance gap with the remarkable deep CF trackers. Equipped with CNN features, CF trackers achieve state-of-the-art tracking accuracy but suffer from a large computational cost.
Methods of feature dimension reduction, such as PCA \cite{CN-DCF}, factorized convolution operator \cite{ECO}, and encoder network \cite{TRACA}, can reduce the feature complexity to some extent. However, these methods have to first extract high-dimensional CNN features from the original heavyweight deep models. In contrast, our method produces a lightweight network offline for efficient feature extraction, which not only naturally reduces the feature dimension but also greatly saves the feature extraction time.

\begin{figure*}
	\centering
	\includegraphics[width=18.3cm]{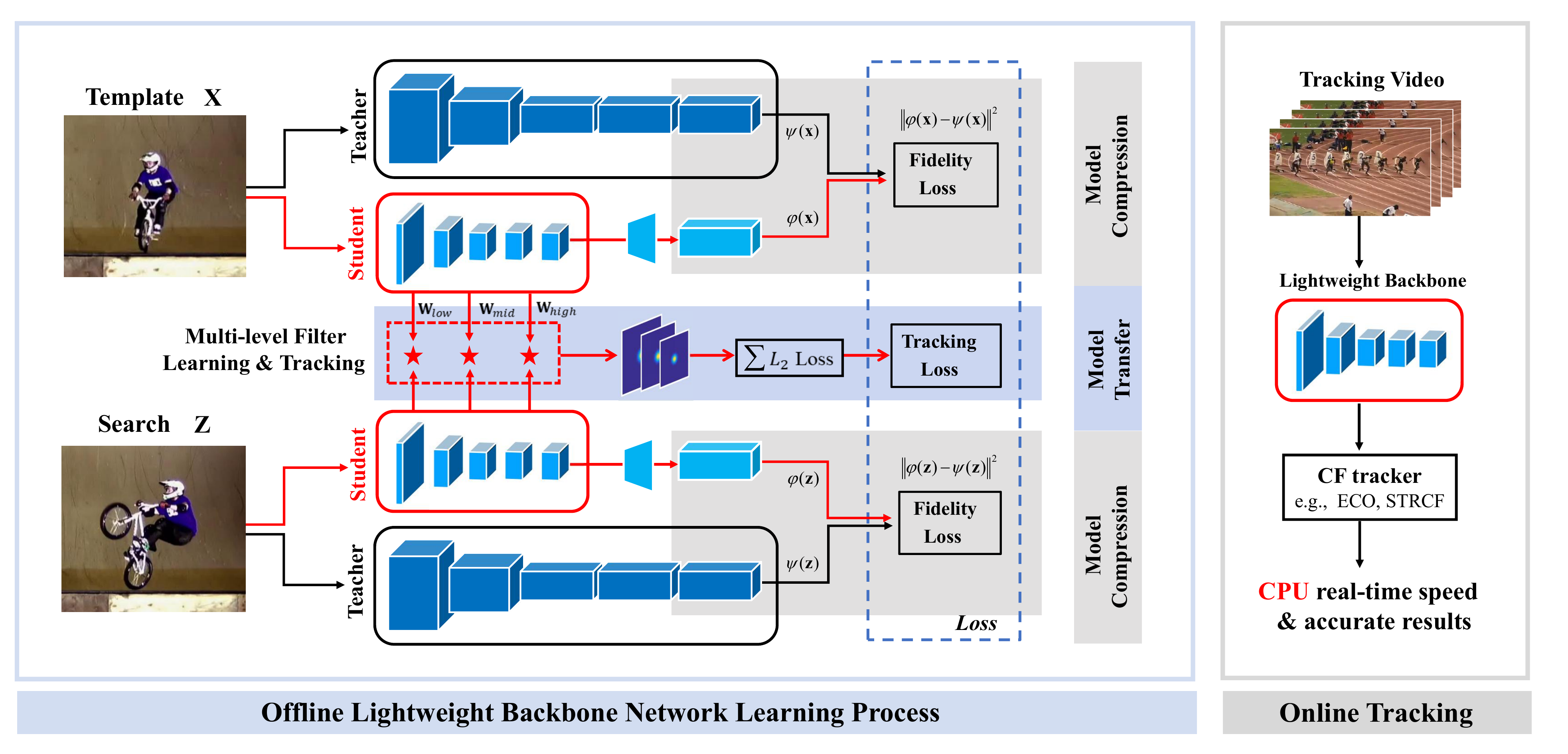}
	\caption{Pipeline of knowledge distillation and online prediction. We learn to offline compress the teacher network by using the proposed fidelity loss and correlation tracking loss. In the online stage, the distilled student network adapts to the target object of each input video sequence and helps track the target object real-time on a single-core CPU.}
	\label{fig:2} \vspace{-0.0in}
\end{figure*}

\subsection{Network Compression}
There are two typical network compression approaches involving model pruning and knowledge distillation. Model pruning \cite{li2016pruning} usually removes unimportant filter weights and utilizes online fine-tuning to recover accuracy.
Knowledge distillation \cite{hinton2015distilling,romero2014fitnets} is based on the observation that a small network has similar representation capability as a large network but is usually harder to train solely \cite{ba2014deep}.
Knowledge distillation \cite{hinton2015distilling,romero2014fitnets} uses a powerful teacher network to guide a smaller student network. The student is forced to mimic the feature representation \cite{romero2014fitnets,ba2014deep} or classification probabilities \cite{hinton2015distilling} of its teacher. However, previous methods usually compress models directly on the same vision task (e.g., image classification).
In contrast, our method not only compresses the deep models but also transfers the objective to the tracking task. Therefore, the distillation and tracking processes are jointly optimized in an end-to-end manner. Unlike existing methods that usually compress the model by $ 4\times $ or $ 8\times $ with limited speed acceleration \cite{romero2014fitnets,hinton2015distilling}, by virtue of collaborative training, we achieve a much larger compression rate of $ 63\times $ while maintaining almost the same tracking accuracy.
In \cite{zhu2017feature_distilled}, the classic knowledge distillation scheme is used to compress off-the-shelf CNN networks in the tracking framework. We note that how to bridge the gap between object recognition and visual tracking is not fully explored. In this work, we propose to simultaneously distill the pretrained networks and narrow the task gap, which helps our method to achieve a much higher compression rate and a real-time speed on CPU. 
To further reduce the model degradation caused by compression, we propose multiple-level knowledge transfer and employ a background-aware online adaption scheme to fine-tune the student network for each sequence.

\section{Revisiting Correlation Tracking} \label{revisit}
A typical CF based tracker \cite{MOSSE,KCF} is trained using an image patch $\bf x $ centered around the target. All of the circular shifts of the target patch ${\bf x}$ are generated as training samples with Gaussian function labels. Considering the feature embedding $ \varphi(\cdot) $, the filter $ \bf w $ can be trained by minimizing the following regularized  regression objective:
\begin{equation}\label{Eq1}
\min_{\bf w}{\left\|\sum_{i=1}^{D}{\varphi_{i}({\bf x})}\star{\bf w}_{i}-{\bf y}\right\|}^{2}+\lambda\sum_{i=1}^{D}{\|{\bf w}_{i}\|}^{2},
\vspace{-0.0in}
\end{equation}
where $ \lambda $ is a regularization parameter, $ D $ is the number of feature channel, $ \star $ denotes the circular correlation and $\bf y $ is the desired Gaussian label. 
The correlation filter on the $ d $-th ($ d\in\{1,\cdots,D\} $) channel can be efficiently learned as follows:
\begin{equation}\label{Eq2}
{\hat{\bf w}_{d}}=\frac{\hat{\bf y}^{*}\odot{\hat{\varphi}_{d}{(\bf x)}}} {{\sum_{i=1}^{D}{\hat{\varphi}_{i}{(\bf x)}}\odot{\hat{\varphi}^{*}_{i}{(\bf x)}} }+\lambda},
\vspace{-0.0in}
\end{equation}
where $\odot$ is the element-wise product, hat notation $\hat{\cdot}$ denotes the Discrete Fourier Transform (DFT) and ${\cdot}^*$ is the complex-conjugate operation.

In the next frame, a search patch $ \bf z $ with the same size of $ {\bf x} $ is cropped out for predicting the target position, and the corresponding response $ \bf r $ is computed by
\begin{equation}\label{Eq3}
{\bf r} = {\cal F}^{-1} \left( \sum^{D}_{i=1}{\hat{\bf w}^{*}_{i}}\odot{\hat{\varphi}_{i}(\bf{z})} \right),
\end{equation}
where $ {\cal F}^{-1}(\cdot) $ is the inverse DFT.
%
Since a higher feature dimension $ D $ implies a larger computation burden, a lightweight feature backbone network not only accelerates feature extraction but also expedites the correlation filter learning (Eq.~\ref{Eq2}) and detection (Eq.~\ref{Eq3}) processes.

In this work, we aim to train a lightweight backbone network for efficient correlation tracking. To verify the effectiveness and generality, we select a baseline and two state-of-the-art CF frameworks as follows:
\begin{itemize}
	\setlength{\parskip}{0pt}
	\item { KCF} \cite{KCF} (in TPAMI 2015) is a plain CF tracker without bells and whistles. We use it to verify the feature representation capability between the teacher and student networks.
	
	\item { ECO} \cite{ECO} (in CVPR 2017) is based on the C-COT \cite{C-COT} tracker and integrates several efficient strategies. ECO adopts the features from VGG-M. We develop the fast version (fECO) using our distilled lightweight model.
	
	\item { STRCF} \cite{STRCF}  (in CVPR 2018) is a CF tracker with a Spatial-Temporal Regularization. STRCF shows impressive performance with hand-crafted features, and DeepSTRCF using VGG-M achieves further improvement but the speed is greatly limited. We implement a fast version, namely fDeepSTRCF, using our model.
\end{itemize}

\section{Proposed Method}\label{sec:proposed approach}

\begin{figure}
	\centering
	\includegraphics[width=6.8cm]{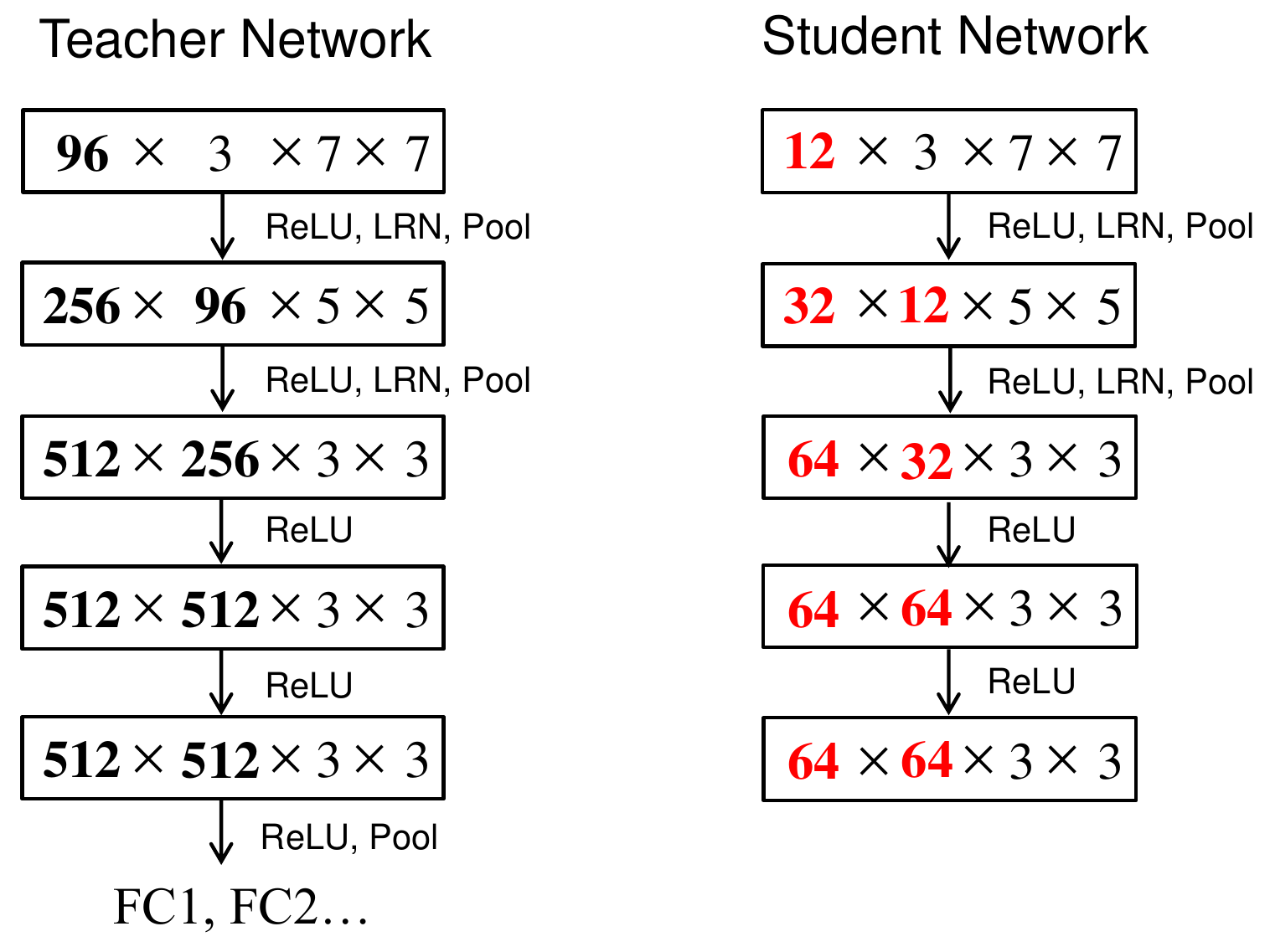}
	\caption{Architecture comparison between the teacher and student networks. The numbers in each convolutional layer indicate the filter number, filter channel, filter width and height, respectively. Notice that the student network reduces 8 times of both filter numbers and channels, which takes around 64 times smaller than the teacher network.}
	\label{fig:3} \vspace{-0.0in}
\end{figure}

Figure~\ref{fig:2} shows an overview of our framework involving offline knowledge distillation and online prediction.
In the offline knowledge distillation step, we use two teacher networks and two shared-weight student networks.
The VGG-M \cite{VGGM} is selected as the teacher network, which is widely used in deep CF trackers \cite{DeepSRDCF,C-COT,ECO,STRCF,DRT}.
We first randomly prune the teacher network to initialize the student network.
Specifically, for one convolutional layer of the teacher network, we randomly prune 7/8 filters in the current layer and the corresponding 7/8 channels in each filter of the next convolutional layer.
The student networks aim to produce similar feature representations of the teacher networks while reducing around 63 times of the model storage.
Figure~\ref{fig:3} shows the detailed architectures of the student and teacher networks where the filter capacity of the student network is 64 times smaller than that of the teacher network in each layer except in the first layer.
%
%
As a result, the teacher network without fully connected layers is 95 MB while our lightweight model is only 1.5 MB. We denote the distilled student network as CF-VGG.

In the following, we first introduce how to offline compress and transfer deep models for efficient tracking in Section \ref{TransferCompress}. Then we present the efficient online adaptation scheme in Section \ref{online adaptation}. 

\subsection{Joint Model Transfer and Compression} \label{TransferCompress}



In the offline training step, we propose two types of losses to simultaneously compress and transfer the teacher network: 1) The fidelity loss ensures the same feature representation capability between the student and teacher networks. 2) The correlation tracking loss transfers the source objective of classification into the target objective of regression for tracking.
The fidelity loss mainly maintains the semantic description in high levels, while the tracking loss learns the similarity (or template matching) to evaluate the minor appearance changes of target objects between frames. By joint training, semantic features can complement the appearance features.
These two losses constitute the final objective function, which is formulated as:
\begin{equation}\label{Eq4}
{\cal L}_{\text{offline}} = {\cal L}_{\text{tracking}} + \lambda {\cal L}_{\text{fidelity}}+\gamma \|\Theta\|^{2},
\end{equation}
where $\lambda$ is a hyper-parameter balancing the influences of these two losses, $\Theta$ denotes the learnable parameters of the student network and the last term is the weight decay. In the following, we present the details of the semantic fidelity loss ${\cal L}_{\text{fidelity}}$ and the correlation tracking loss ${\cal L}_{\text{tracking}}$.


\subsubsection{Semantic Fidelity Loss}\label{fidelity loss}



Once we initialize the student network by filter pruning, the feature dimensions of the student and teacher networks are different. We use a 1$\times$1 fully convolutional operation to match their feature dimension. Given a target patch $\bf x$ and a search patch $\bf z$, the features from the student network and the teacher network should be as similar as possible. We propose a fidelity loss to measure the feature differences. Formally, we define the fidelity loss as:
\begin{equation}\label{Eq5}
\begin{aligned}
{\cal L}_{\text{fidelity}} & = {\cal L}_{\text{target}} + {\cal L}_{\text{search}}\\
& = \|\varphi({\bf x}) - \psi({\bf x}) \|^{2} + \|\varphi({\bf z}) - \psi({\bf z}) \|^{2},
\end{aligned}
\end{equation}
where $ \varphi(\cdot) $ represents the trainable feature embedding of the student network (its notation $ \Theta $ is omitted for clarity), and $ \psi(\cdot) $ is the fixed embedding of the teacher network.


\subsubsection{Correlation Tracking Loss}\label{correlation tracking loss}

In addition to the fidelity loss, we propose the correlation tracking loss to modify the objective of the student network from classification to regression.
We feed the search patch and the target patch into the student network to obtain their features and use a CF to model the response map regression.
The circular correlation can be computed in the Fourier domain with a closed-form solution \cite{KCF,DSST} and the backward formulas can also be efficiently derived.
The corresponding loss function is the $ L_{2} $ distance between the correlation response map ${\bf r}$ and the groundtruth label ${\bf g}$ as follows:
\begin{equation}\label{Eq6}
\begin{aligned}
{\cal L}_{\text{tracking}} &= \|{\bf r}-{\bf g}\|^{2},\\
s.t.~~~~~~~~{\bf r} &= {\cal F}^{-1} \left({\hat{\bf w}^{*}}\odot{\hat{\varphi}(\bf{z})} \right), \\
{\hat{\bf w}}&=\frac{\hat{\bf y}^{*}\odot{\hat{\varphi}{(\bf{x})}}} {{{\hat{\varphi}{(\bf{x})}}\odot{\hat{\varphi}^{*}{(\bf{x})}} }+\lambda},
\end{aligned}
\end{equation}
where $ {\bf g} $ is a Gaussian map centered at the annotated target location. For clarity, in comparison with Eq.~\ref{Eq2}, we omit the feature dimension $ D $ in Eq.~\ref{Eq6} and the subsequent equations. The back-propagation of the above loss with respect to $ \varphi(\bf x) $ and $ \varphi{\bf (z)} $ are given by Eq.~\ref{Eq7} below. Interested readers can refer to \cite{CFNet, DCFNet} for more details.
\begin{equation}\label{Eq7}
\begin{aligned}
\nabla_{\varphi(\bf x)}{\cal L} &= {\cal F}^{-1}\left( \nabla_{\hat{\varphi}^{*}(\bf x)}{\cal L} + (\nabla_{ \hat{\varphi}(\bf x)}{\cal L})^{*} \right),\\
\nabla_{\varphi(\bf z)}{\cal L} &= {\cal F}^{-1}\left( \nabla_{\hat{\varphi}^{*}(\bf z)}{\cal L} \right).
\end{aligned}
\end{equation}

\begin{figure}
	\centering
	\includegraphics[width=7.5cm]{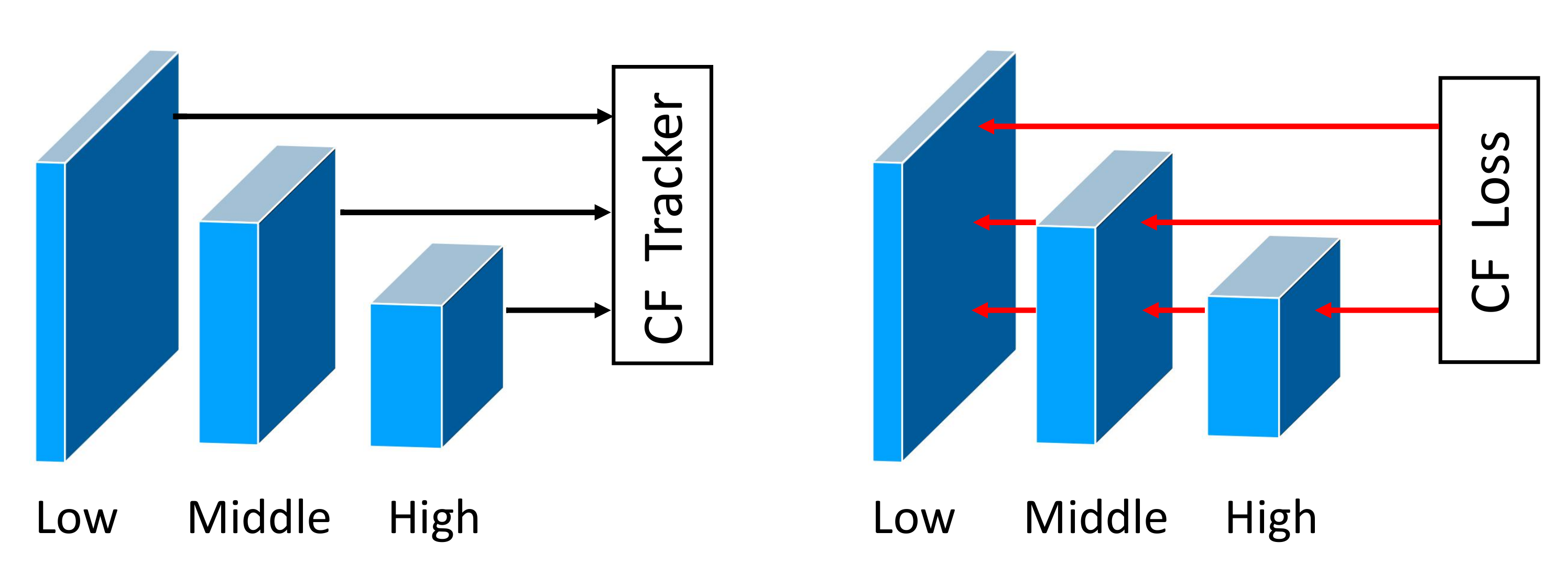}
	\caption{Existing deep CF trackers learn correlation filters directly on multi-layer CNN features (left). Our framework aims at distilling a lightweight CNN backbone using back propagation to fine-tune multiple layers (right). }
	\label{fig:4}
	\vspace{-0.0in}
\end{figure}

Furthermore, we perform the model transfer with multiple-level feature representations. Unlike existing deep CF trackers \cite{HCF,HDT,C-COT,ECO} that simply integrate multiple CNN layers with empirical or learnable weights to boost the performance (see Figure~\ref{fig:4}), we separately apply the trainable constraint to multiple CNN layers to fine-tune the student network. This helps the student network not only fit the correlation tracking task better but also maintain a richer representation capability than only using the features from the last CNN layer (see more experiments in Section~\ref{ablation study}). In this work, we take the first, second and last convolutional layers before their pooling operations as the low, middle and high-level feature representations, respectively. The final tracking loss is formulated as:
\begin{equation}\label{Eq8}
\begin{aligned}
{\cal L}_{\text{tracking}} 
&= \sum_{l}\|{\bf r}_{l}-{\bf g}_{l} \|^{2}, \\
s.t.~~~~{\bf r}_{l} &= {\cal F}^{-1} \left( {\hat{\bf w}^{*}_{l}}\odot{\hat{\varphi}_{l}({\bf z})} \right),\\
l~ &\in \{\text{high, middle, low}\},
\end{aligned}
\end{equation}
where $ l $ means the  index  of the feature representation level. $ {\bf g}_{l} $ contains the groundtruth labels, which are all Gaussian maps but with different spatial sizes. $ \varphi_{l}(\cdot) $ denotes the feature embedding of the student network on the $ l $-th level. The CFs with different levels of features (i.e., $ {\bf w}_{l}$) are learned using Eq.~\ref{Eq2}.

\subsection{Background-Aware Online Adaptation} \label{online adaptation}

\begin{figure}
	\centering
	\includegraphics[width=8.6cm]{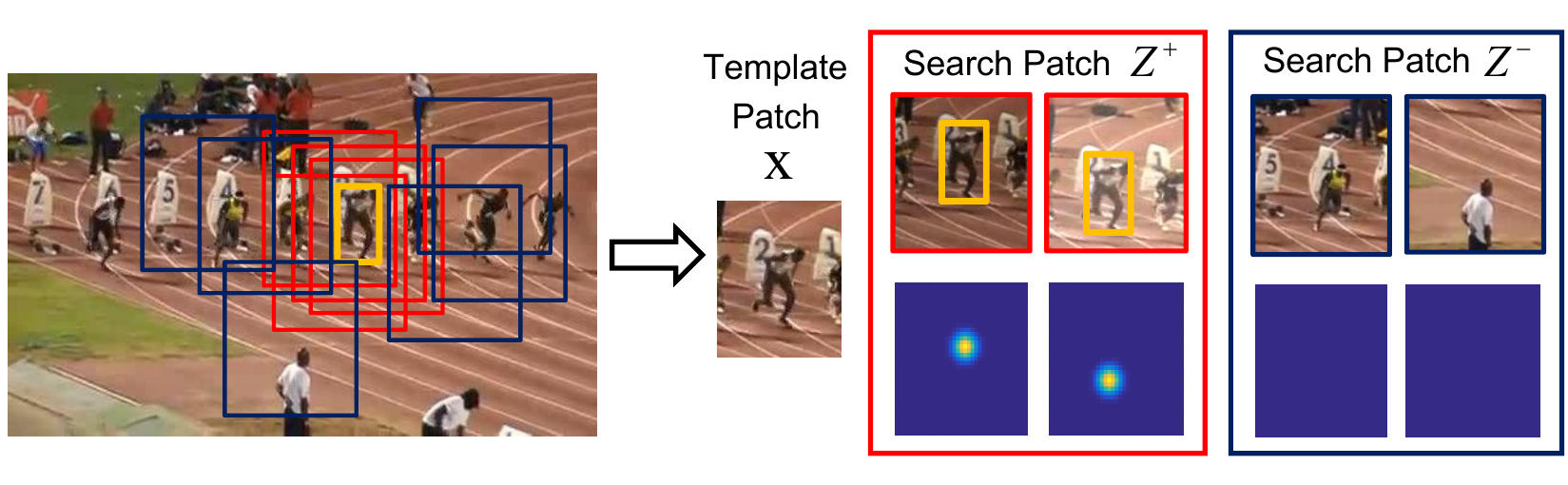}
	\caption{Illustration of sample generation for background-aware online adaptation. Given the first frame, the template {\bf x} is cropped centered at the target position. The foreground patches $ {\bf z}^{+} $ contain the target. We augment the foreground patches for training. The background patches $ {\bf z}^{-} $ do not include the target and their corresponding labels are set to zero.}
	\label{fig:5} 
\end{figure}

The offline distillation decreases the network capacity while preserving the feature representation.
In the tracking scenarios, objects belonging to the same category may be labeled differently according to the first frame annotations.
Figure~\ref{fig:5} shows an example where only one athlete is positively labeled while the remaining are labeled as negative.
In order to increase the feature discrimination, we online fine-tune the student network using the annotations in the first frame.
Our idea is motivated by the context-aware correlation filter (CACF) \cite{Context-AwareCorrelationFilter} that regresses hard negative samples $ {\bf x}^{-} $ to the negative labels.
These hard negative samples do not overlap with the target object.
In CACF \cite{Context-AwareCorrelationFilter}, the context-aware information is learned through:
\begin{equation}\label{Eq9}
\min_{\bf w}{\|\varphi({\bf x}^{+})\star{\bf w}-{\bf y}\|}^{2}+\lambda_{1}{\|{\bf w}\|}^{2}+\lambda_{2}\sum_{i=1}^{k}{\|\varphi({\bf x}^{-}_{i})\star{\bf w}\|}^{2},
\end{equation}
where ${\bf x^+}$ is the positive training sample including the target and ${\bf x}_{i}^{-}$ collects the negative samples that do not overlap with the target region.
%
Given pretrained deep features, the CACF method enhances the filter-level discriminative capability. However, in our work, we explore the background-aware information in model training to boost the feature-level representation.

In the offline distillation step, all the training samples contain the target object, which helps discriminate the target from the background in a limited neighborhood.
During online fine-tuning, we incorporate more negative samples to help the student network better distinguish the target from the background where hard negative objects may exist.
To this end, we crop both positive and negative samples online, as shown in Figure~\ref{fig:5}.
For positive samples, we augment them through randomly flipping, shifting, increasing blur, and changing the illumination.
Finally, the target patch $ {\bf x} $ is fed into the template branch. Positive $ {\bf z}^{+} $ and negative $ {\bf z}^{-} $ search samples are fed into the search branch in Figure~\ref{fig:2}. For online adaptation, we jointly exploit the multi-level transfer and background-aware formulation. The online tracking loss $ {\cal L}^{'}_{\text{tracking}} $ is as follows:
\begin{equation}\label{Eq10}
\begin{aligned}
{\cal L}^{'}_{\text{tracking}} 
& = \sum_{l}\left(\|{\bf r}^{+}_{l}-{\bf g}_{l} \|^{2} + \|{\bf r}^{-}_{l}\|^{2}\right), \\
s.t.~~~~{\bf r}^{+}_{l} &= {\cal F}^{-1} \left( {\hat{\bf w}^{*}_{l}}\odot{\hat{\varphi}_{l}({\bf z^{+}})} \right),\\
{\bf r}^{-}_{l} &= {\cal F}^{-1} \left( {\hat{\bf w}^{*}_{l}}\odot{\hat{\varphi}_{l}({\bf z^{-}})} \right),\\
l~ &\in \{\text{high, middle, low}\},
\end{aligned}
\end{equation}
where $ + $ and $ - $ on the label $ \bf r $ and search patch $ \bf z $ denote the positive and negative annotations, respectively. As for the fidelity loss, since we still want the student network to mimic its teacher on the current video, $ {\cal L}_{\text{fidelity}} $ is kept the same as in Eq.~\ref{Eq5}. 
The online fine-tuning is only performed on the initial frame and its loss is given as follows:
\begin{equation}\label{Eq11}
{\cal L}_{\text{online}} = {\cal L}^{'}_{\text{tracking}} + \lambda {\cal L}_{\text{fidelity}}+\gamma \|\Theta\|^{2}.
\end{equation}

\subsection{Efficient Online Correlation Tracking}

After we have the distilled lightweight backbone, we remove the additionally added 1$ \times $1 convolutional kernel, and take the output of the remaining convolutional layers to facilitate existing CF frameworks for online tracking. We select three representative methods (i.e., KCF \cite{KCF}, ECO \cite{ECO}, and STRCF \cite{STRCF}) as introduced in Section \ref{revisit}.

\section{Experiments}\label{sec:exp}
In this section, we first illustrate the implementation details and the evaluation configurations. Then we conduct an ablation study to demonstrate the effectiveness of our method. Finally, we compare with state-of-the-art trackers.

\subsection{Experimental Details}

{\flushleft \bf Implementation Details.} We use the videos for object detection from the ImageNet Large Scale Visual Recognition Challenge (ILSVRC 2015) \cite{ILSVRC2015} dataset to offline distill the student network. During training, we use the stochastic gradient descent (SGD) solver and set the momentum and weight decay as 0.9 and 0.005, respectively. We train the network for 50 epochs with a learning rate exponentially decreased from $10^{-2}$ to $10^{-5}$. The multi-task weighting parameter $\lambda$ in Eq. \ref{Eq4} and Eq. \ref{Eq11} is set to $ 10^{-5} $. In the online adaptation stage, we fine-tune the student network for only 8 iterations using the samples from the first frame. In each iteration, we crop 32 positive and negative samples as shown in Figure~\ref{fig:5}. We implement our method using MatConvNet \cite{Vedaldi2014MatConvNet} on a PC with a 4GHz CPU and an Nvidia GTX 1080TI GPU. The source code will be available at:  \url{https://github.com/594422814/CF-VGG.git}


{\flushleft \bf Benchmarks and Evaluation Metrics.} We evaluate our tracker on the OTB-2013 \cite{OTB-2013}, OTB-2015 \cite{OTB-2015}, and Temple-Color \cite{TempleColor128} datasets, which contain 50, 100 and 128 challenging videos, respectively. 
We report the overlap success plots on these datasets using one-pass evaluation (OPE) \cite{OTB-2013,OTB-2015} and take the area-under-curve (AUC) scores to evaluate the performance. 
In addition, we evaluate our tracker on the VOT-2016 \cite{VOT2016} and VOT-2017 \cite{VOT2017} datasets. The performance is measured by two independent metrics: accuracy (average overlap during successful tracking) and robustness (reset rate).

\subsection{Ablation Study}\label{ablation study}

We evaluate the effectiveness of the components of the proposed algorithm in terms of computational efficiency, tracking accuracy, and model representation capability.

{\flushleft \bf Efficiency.} Table~\ref{table:table1} compares the efficiency and model size of our CF-VGG with the original teacher network VGG-M. These two networks are integrated into the state-of-the-art CF trackers ECO and DeepSTRCF. We observe that it takes around 76 ms for the VGG-M network to extract features on the CPU, which is 8 times slower than that using our distilled CF-VGG network. The distilled deep features accelerate the ECO and DeepSTRCF trackers and are more than 5 times faster on the CPU.
The improved fECO and fDeepSTRCF trackers take 27 FPS and 20 FPS vs. their original speed 5 FPS and 3 FPS, respectively.



\begin{figure}
	\centering
	\includegraphics[width=8.9cm]{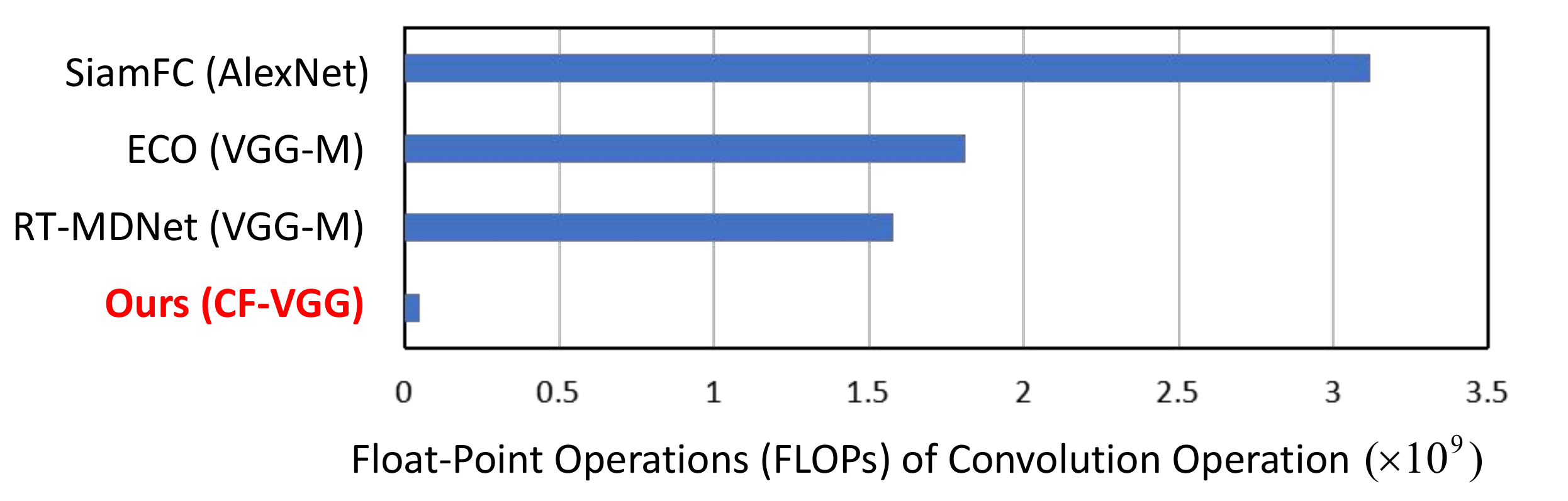}
	\caption{Model computation complexity comparison. Our proposed lightweight CF-VGG produces much fewer FLOPs, which guarantees the CPU real-time correlation tracking.} \label{fig:flop} \vspace{-0.0in}
\end{figure}

In addition to the comparison with CF trackers using VGG-M, we further analyze some other representative real-time trackers. Figure \ref{fig:flop} shows the comparison results of some widely adopted backbones on FLOPs metric (only feature extraction part). The number of float-point operations (FLOPs) of the convolutional layer is calculated as follows,
\begin{equation}\label{FLOP}
\text{FLOPs} = (C_{\text{in}}K^{2}+1)HWC_{\text{out}},
\end{equation}
where $ C_{\text{in}} $ is the input feature map channel, $ K $ is the kernel width (assumed to be symmetric), +1 means the computation of bias operation, and $ H $, $ W $ and $ C_{\text{out}} $ are the  height, width and channel number of the output feature maps, respectively. In Table \ref{table:feature size}, we exhibit the feature map sizes and feature channels of different backbone networks. The AlexNet backbone is typically used in Siamese trackers \cite{SiamFc,SASiam} and the VGG-M network is widely adopted in classification based trackers \cite{MDNet,VITAL,DATnips} and CF trackers \cite{ECO,STRCF}.  After computing the FLOPs of different backbone networks via Eq.~\ref{FLOP}, we can observe that our tiny model is extremely efficient than modern off-the-shelf models.
The FLOPs of the feature extractor in SiamFC and ECO are $ 3.12\times10^{ 9} $ and $1.82\times10^{9} $ while ours is only $ 4.79\times10^{7} $, as shown in Figure~\ref{fig:flop}.

\setlength{\tabcolsep}{2pt}
\begin{table}
	\scriptsize
	\begin{center}
		\caption{Computation comparison between the ECO/DeepSTRCF trackers and our improved versions on the OTB-2013 dataset. We use float-point operations (FLOPs) of convolution operation to measure the computational complexity, where B indicates billion. In practice, the actual speedup ratio is much slower than FLOPs.} \label{table:table1}	
		\vspace{+0.08in}
		\begin{tabular*}{8.2 cm} {@{\extracolsep{\fill}}lcccccc}
			\hline
			&Backbone & Model & Model & CPU Feature  & CPU & GPU    \\
			&Model& Size &FLOPs & Extraction  & FPS & FPS    \\
			\hline
			~ECO \cite{ECO} &VGG-M \cite{VGGM} & 95 MB & 1.82 B &76 ms   &5 &9 \\
			~fECO  &CF-VGG & 1.5 MB & 0.048 B &9 ms  &27 &$ > $48 \\
			\hline
			~DeepSTRCF \cite{STRCF}   &VGG-M \cite{VGGM} & 95 MB &1.82 B &76 ms  &3 &5 \\
			~fDeepSTRCF  &CF-VGG & 1.5 MB &0.048 B &9 ms  &20 &$ > $35 \\
			\hline
		\end{tabular*}
	\end{center}
	\vspace{-0.0in}
\end{table}

\setlength{\tabcolsep}{2pt}
\begin{table}
	\scriptsize
	\begin{center}
		\caption{Comparison of the feature map sizes and feature channels of diferent networks including AlexNet \cite{Alexnet}, VGG-M \cite{VGGM} and our CF-VGG.} \label{table:feature size}
		\vspace{+0in} 
		\begin{tabular*}{8.4 cm} {@{\extracolsep{\fill}}lccc}
			\hline
			& AlexNet (SiamFC \cite{SiamFc}) &VGG-M (ECO \cite{ECO}) &CF-VGG (fECO) \\
			\hline
			~Input  &$ 255\times255\times3 $ &$ 224\times224\times3 $ &$ 224\times224\times3 $  \\
			~Conv1 &$ 123\times123\times96 $ &$ 112\times112\times96 $ &$ 112\times112\times{\bf 12} $ \\
			~Pool1 &$ 61\times61\times96 $ &$ 56\times56\times96 $ &$ 56\times56\times12 $ \\
			~Conv2 &$ 57\times57\times256 $  &$ 28\times28\times256 $ &$ 28\times28\times{\bf 32} $ \\
			~Pool2 &$ 28\times28\times256 $  &$ 14\times14\times256 $ &$ 14\times14\times32 $ \\
			~Conv3 &$ 26\times26\times192 $  &$ 14\times14\times512 $ &$ 14\times14\times{\bf 64} $ \\
			~Conv4 &$ 24\times24\times192 $  &$ 14\times14\times512 $ &$ 14\times14\times{\bf 64} $ \\
			~Conv5 &$ 22\times22\times128 $  &$ 14\times14\times512 $ &$ 14\times14\times{\bf 64} $ \\
			\hline
		\end{tabular*}
	\end{center}
\end{table}

The Siamese trackers \cite{SiamFc,StructSiam,SASiam,Dsiam} adopt AlexNet-like \cite{Alexnet} fully-convolutional networks to predict target location in an end-to-end manner. Their tracking speed can be significantly accelerated to over 80 FPS by a powerful GPU because the fully-convolutional structure adequately exploits the GPU device. However, on a single CPU, the Siamese trackers are unlikely to achieve real-time performance \cite{EAST}, whereas our improved CF trackers can. The recent real-time MDNet tracker \cite{RTMDNet} modifies the first three convolutional layers of VGG-M and uses ROI Align for efficient binary classification. However, its backbone network still produces high FLOPs and the further online fine-tune prevents its CPU real-time performance. 
For CF trackers, only the deep feature extraction process benefits from GPU and the tracking part just uses CPU even without optimization. Besides, ECO \cite{ECO} and STRCF \cite{STRCF} methods use a  time-consuming alternating direction method of multipliers (ADMM) or Conjugate Gradient (CG) for online algorithm optimization. Thus, existing speed comparison that does not distinguish CPU and GPU environments is not very fair. With only CPU, the Siamese tracker (e.g., SiamFC) of more than 80 FPS cannot achieve real-time speed \cite{EAST} but ours can make it, which already proves the efficiency of CF trackers using our tiny model. 

%

{\flushleft \bf Compression Ratio.} In this work, we compress the off-the-shelf model by an extremely high ratio of about 63$ \times $. In Table \ref{table:ratio}, we evaluate the performance, model size and CPU speed under different network compression ratios. The 1$ \times $ compression ratio means that the network is not pruned, but still fine-tuned by the fidelity loss and correlation tracking loss. It slightly outperforms the teacher model, which shows that our joint training scheme is effective and the tracking loss slightly fine-tunes the uncompressed model. To achieve CPU real-time speed, we choose the compression ratio of 64$ \times $. Except for the better efficiency, by adopting our lightweight model, the required storage room is also greatly saved (our 1.5 MB vs. original 95 MB).

\setlength{\tabcolsep}{2pt}
\begin{table}
	\scriptsize
	\begin{center}
		\caption{Performance and speed analysis on different compression ratios. The baseline tracker is ECO, and is evaluated on the  OTB-2015 benchmark \cite{OTB-2015} using AUC metric. To achieve both satisfying CPU real-time efficiency and performance, we choose the compression rate of 64$ \times $.} \label{table:ratio}
		\vspace{+0.in} 
		\begin{tabular*}{8.4 cm} {@{\extracolsep{\fill}}lcccccc}
			\hline
			~Compression Ratio  &Baseline &1$ \times $ &16$ \times $ &32$ \times $ &64$ \times $ &96$ \times $ \\
			\hline
			~Model Size (MB) &95 MB &95 MB &5.8 MB &2.9 MB &1.5 MB &0.99 MB \\
			~AUC Socre (\%)  &69.4  &69.6 &69.0 &68.5 &68.2 &66.9\\
			~CPU Speed (FPS)  &5  &5 &9 &16 &27 & 35\\
			\hline
		\end{tabular*}
	\end{center}
\end{table}


{\flushleft \bf Tracking Accuracy.} In Table~\ref{table:table2}, we show the performance evaluation results using different configurations to distill the student network. 
To obtain a tiny model, an optional choice is directly training a tiny CF-VGG from scratch using classification loss following VGG-M, but its performance is unsatisfied since it may not suit the tracking task. 
In contrast, we propose to jointly compress and transfer a teacher network. When using only the fidelity loss (i.e., shown as ``only fidelity loss"), the tracking accuracy decreases by 4$ \sim $5\% for ECO. Meanwhile, using only tracking loss decreases the accuracy by 3$ \sim $4\% as well. However, equipped with both the fidelity and tracking losses, we significantly improve the performance, which means the high-level semantic features can complement the multi-level appearance features trained via tracking loss.
When integrated into DeepSTRCF, we find that the improved fDeepSTRCF tracker achieves higher accuracy on both the OTB-2013 and OTB-2015 datasets. Our performance slightly decreases on the Temple-Color dataset.
Finally, with online adaptation (i.e., ``offline + online''), the trackers show slightly better results and the performance gap is only about 1$ \sim $2\% compared to the baselines with uncompressed deep features.
In addition, our improved fECO and fDeepSTRCF trackers achieve much higher performance than ECOhc and STRCF, which both use hand-crafted features.

For tracking speed computation, we do not include the initial adaptation time, which will slightly reduce the average speed (about 3$ \sim $5 FPS on a CPU). However, it is worth mentioning that the offline pretrained CF-VGG model already works well even without online adaptation.



\setlength{\tabcolsep}{2pt}
\begin{table}
	\scriptsize
	\begin{center}
		\caption{Comparison of tracking accuracy under different training configurations. We report AUC scores on the OTB-2013 \cite{OTB-2013}, OTB-2015 \cite{OTB-2015}, and Temple-Color \cite{TempleColor128} datasets. The values in brackets denote the performance gap compared with the corresponding baseline with uncompressed deep model.}
		\label{table:table2}	
		\vspace{0.02in}
		\begin{tabular*}{8.6 cm} {@{\extracolsep{\fill}}lccc}
			\hline
			~Trackers of different variations  & OTB-2013  & OTB-2015 & TC-128~     \\
			\hline
			~ECO (baseline)  &71.0 &69.4 & 60.3~   \\
			~ECOhc (hand-crafted feature)  &65.6 &64.6 &54.7~ \\
			~fECO (pretrained tiny model)  &66.0 (-5.0) &65.5 (-3.9) & 55.4 (-4.9)~ \\
			~fECO (only fidelity loss)  &66.1 (-4.9) &65.1 (-4.3) & 55.0 (-5.3)~ \\
			~fECO (only tracking loss, single-scale)  &65.2 (-5.8) &64.6 (-4.8) & 54.6 (-5.7)~ \\
			~fECO (only tracking loss, multi-scale)  &66.3 (-4.7) &65.9 (-3.5) & 55.9 (-4.4)~ \\
			~fECO (fidelity + multi-scale tracking)  &68.4 (-2.6)  &67.9 (-1.5) & 57.4 (-2.9)~    \\
			~fECO (offline + online fine-tune)   &68.5 (-2.5)  &68.2 (-1.2) & 57.4 (-2.9)~     \\
			\hline
			~DeepSTRCF (baseline)  &69.2 &68.5 & 59.9~   \\
			~STRCF (hand-crafted feature)  &66.5 &64.8 & 54.9~ \\
			~fDeepSTRCF (pretrained tiny model)  &65.1 (-4.1) &65.2 (-3.3) &55.2 (-4.7)~ \\
			~fDeepSTRCF (only fidelity loss)  &65.6 (-3.6) &65.5 (-3.0) &55.1 (-4.8)~ \\
			~fDeepSTRCF (only tracking loss, single-scale)  &65.7 (-3.5)& 65.4 (-3.1) &54.8 (-5.1)~ \\
			~fDeepSTRCF (only tracking loss, multi-scale)  &66.9 (-2.3)& 66.0 (-2.5) &55.5 (-4.4)~ \\
			~fDeepSTRCF (fidelity + multi-scale tracking)  &69.4 (+0.2)  &67.8 (-0.7) & 56.9 (-3.0)~     \\
			~fDeepSTRCF (offline + online fine-tune)    &70.3 (+1.1)  &68.6 (+0.1) &57.3 (-2.6)~   \\
			\hline
		\end{tabular*}
	\end{center}
	\vspace{-0.0in}
\end{table}

\setlength{\tabcolsep}{2pt}
\begin{table}
	\scriptsize
	\begin{center}
		\caption{Feature representation capability comparison between VGG-M and our compressed model. We present AUC scores on the OTB-2015 \cite{OTB-2015} dataset. Our fKCF achieves comparable performance on each single feature layer with its teacher. } \label{table:table3}	
		\vspace{+0.08in}
		\begin{tabular*}{8.2 cm} {@{\extracolsep{\fill}}lcccccc}
			\hline
			&Backbone & Model & Low-level & Middle-level & High-level & CPU  \\
			&Model & Size & Conv 1 & Conv 2 & Conv 5 & FPS    \\
			\hline
			~KCF &VGG-M \cite{VGGM}   & 95 MB &48.0 &50.6 &49.2 &6 \\
			~fKCF &CF-VGG  & 1.5 MB &46.8 (-1.2) &51.0 (+0.4) &47.1 (-2.1) &48\\		
			\hline
		\end{tabular*}
	\end{center}
	\vspace{-0.1in}
\end{table}



{\flushleft \bf Model Representation Capability.} The state-of-the-art CF trackers (i.e., ECO and DeepSTRCF) employ spatial regularization to reduce boundary effects in learning correlation filters. This may leave the concern that how likely the compressed CF-VGG model can maintain the presentation capability of deep models. To demonstrate the effectiveness of CF-VGG, we use the baseline KCF method \cite{KCF} to evaluate the performance on each single feature level without bells and whistles. Table~\ref{table:table3} shows that KCF with CF-VGG exhibits comparable performance with the original teacher network. This clearly indicates that the distilled student network almost maintains the same feature representation capability even though its model size is 63 times smaller than its teacher network.


\begin{figure}
	\centering
	\includegraphics[width=4.3cm]{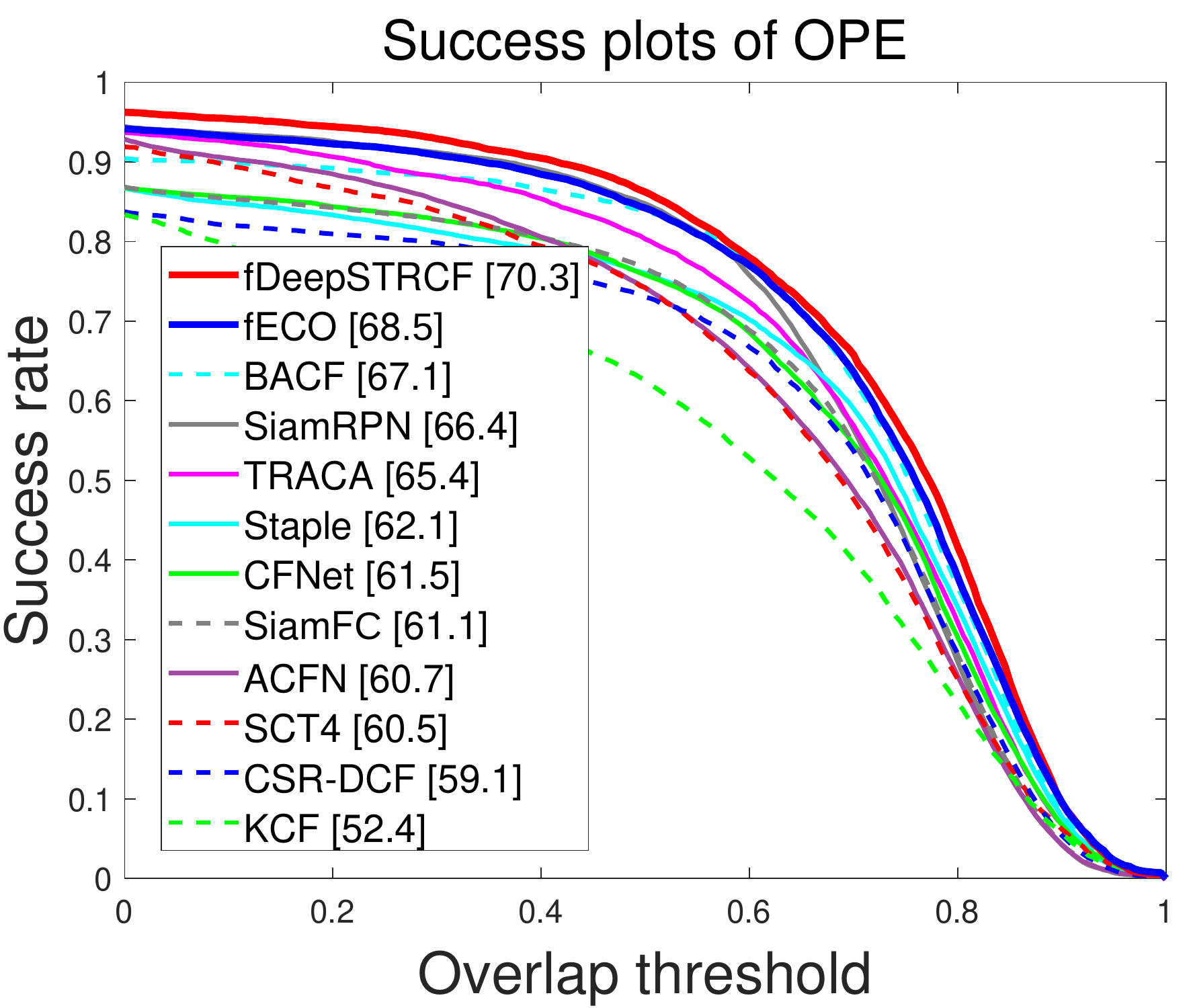}
	\includegraphics[width=4.3cm]{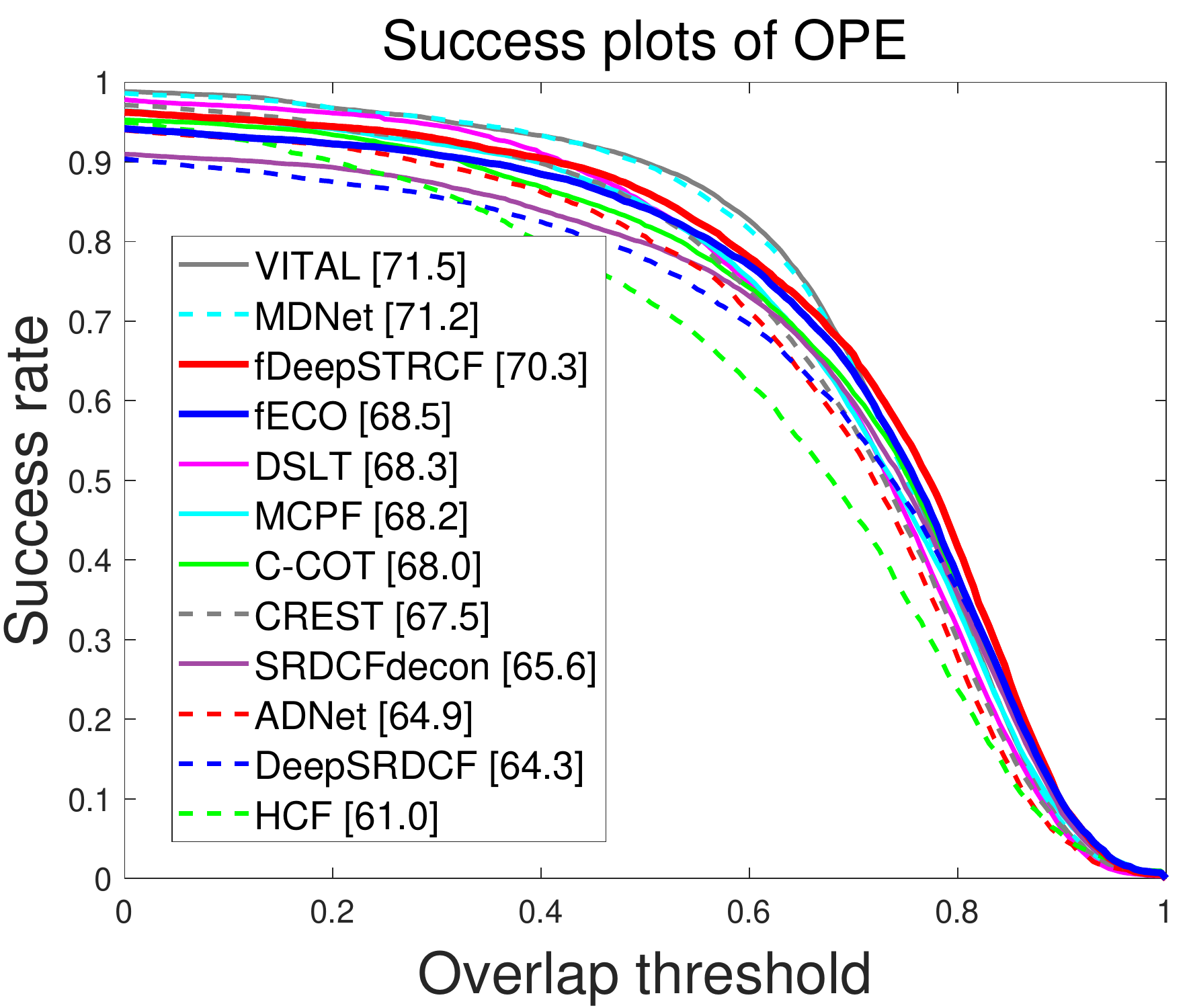}
	\caption{Success plots of real-time trackers (left) and non-realtime trackers (right) on the OTB-2013 \cite{OTB-2013} dataset. In the legend, we show the area-under-curve (AUC) score.} \label{fig:6} \vspace{-0.0in}
\end{figure}

\begin{figure}
	\centering
	\includegraphics[width=4.3cm]{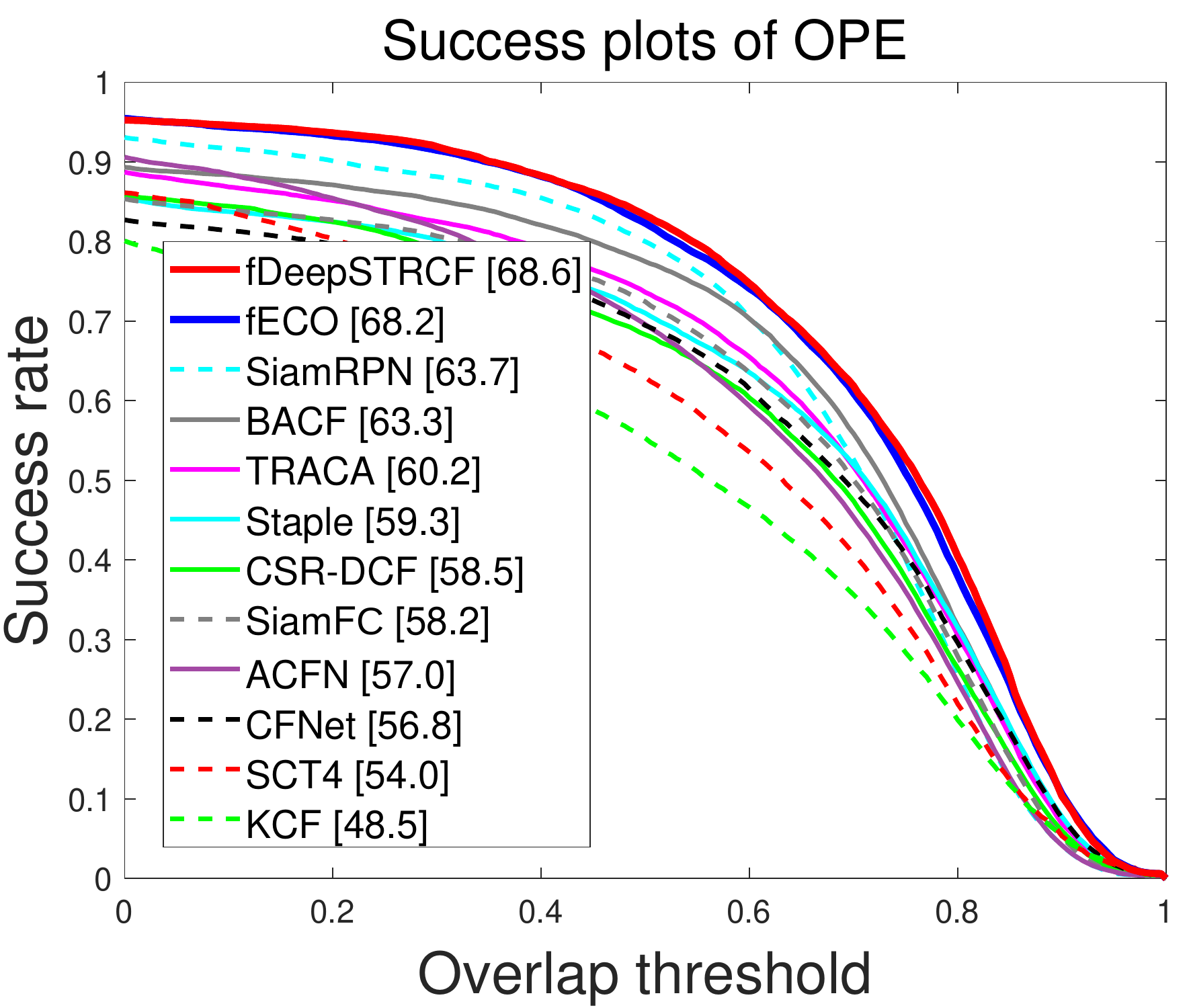}
	\includegraphics[width=4.3cm]{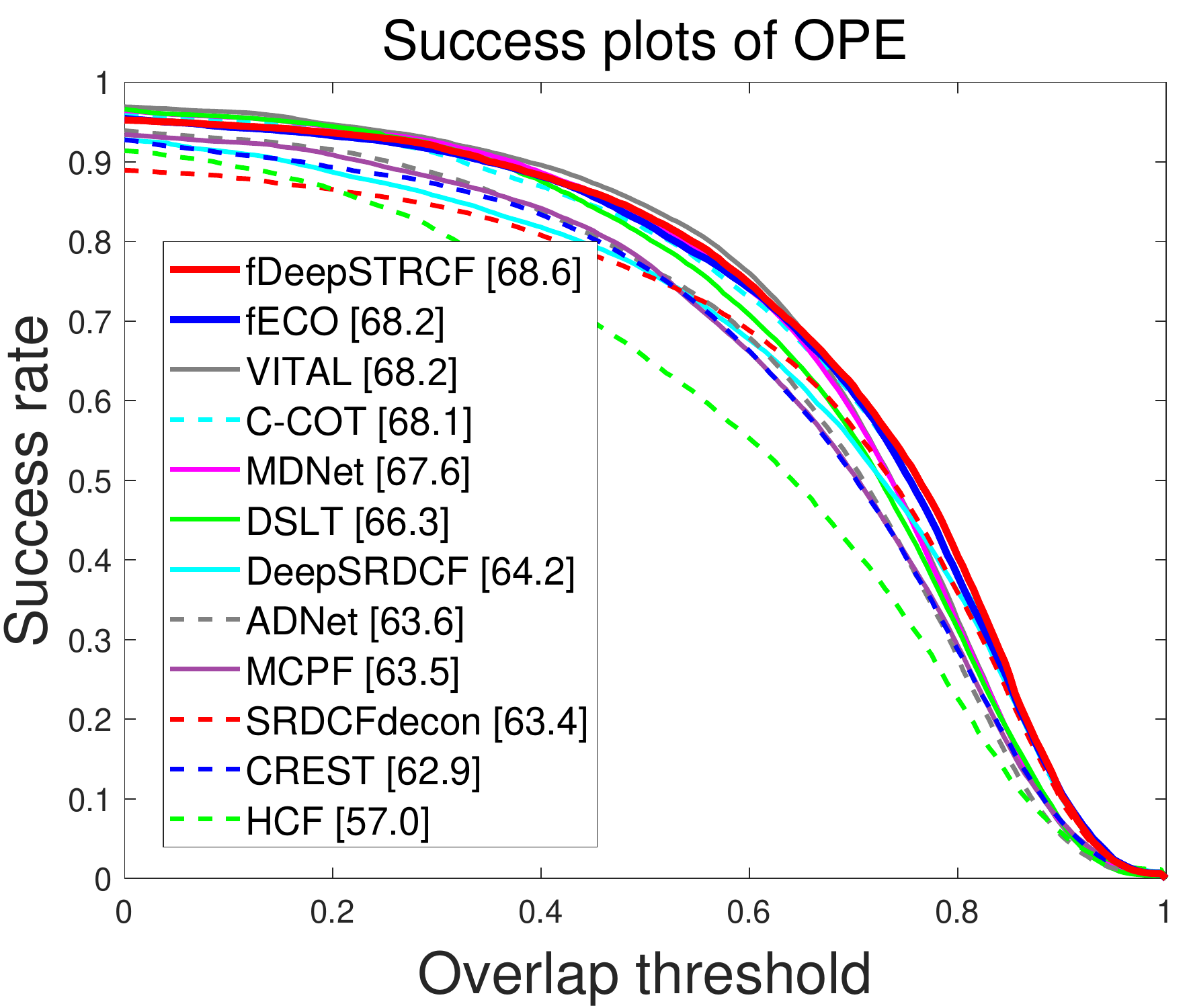}
	\caption{Success plots of real-time trackers (left) and non-realtime trackers (right) on the OTB-2015 \cite{OTB-2015} dataset. Our trackers obviously surpass other real-time methods and even outperform most non-realtime deep trackers.} \label{fig:7} \vspace{-0.0in}
\end{figure}

\subsection{Comparison with State-of-the-arts} \label{state-of-the-art comparison}
We compare our fECO and fDeepSTRCF with 20 state-of-the-art trackers, which are mainly categorized as real-time trackers and non-realtime trackers.

\begin{itemize}
	\item { Real-time Trackers:} For comprehensive comparison, we collect recent high-performance real-time trackers including TRACA \cite{TRACA} (100 FPS), SiamRPN \cite{SiamRPN} (160 FPS), BACF \cite{BACF} (35 FPS), CFNet \cite{CFNet} (65 FPS), CSR-DCF \cite{CSR-DCF} (15 FPS), ACFN \cite{ACFN} (15 FPS), SiamFC \cite{SiamFc} (86 FPS), Staple \cite{Staple} (70 FPS), SCT4 \cite{SCT} (50 FPS), and KCF \cite{KCF} (270 FPS). It should be noted that some of these trackers require GPU to achieve high speed (e.g., TRACA, SiamRPN, CFNet, ACFN, and SiamFC). In contrast, our methods are free of such requirement.
	
	\item { Non-realtime Trackers:} We compare with high accuracy trackers including VITAL \cite{VITAL} (1.5 FPS), DSLT \cite{DSLT} (5 FPS), CREST \cite{CREST} (3 FPS), MCPF \cite{MCPF} (2 FPS), ADNet \cite{ADNet} (1 FPS), C-COT \cite{C-COT} (0.3 FPS), MDNet \cite{MDNet} (1 FPS), SRDCFdecon \cite{SRDCFdecon} (3 FPS), DeepSRDCF \cite{DeepSRDCF} ($ < $1 FPS), and HCF \cite{HCF} (12 FPS). Among these trackers, only C-COT and SRDCFdecon are tested on CPU and all the other trackers rely on a high-end GPU. Although these trackers achieve state-of-the-art performance on the benchmarks, their computational load limits the practical usage. In the following experiments, we will show that our CPU real-time methods still outperform most of them.
\end{itemize}

{\flushleft \bf OTB-2013 Dataset.} On the OTB-2013 benchmark, our fECO and fDeepSTRCF achieve the AUC scores of 68.7\% and 70.5\%, respectively. Figure~\ref{fig:6} (left) shows that our trackers perform better over other real-time trackers such as the recent BACF \cite{BACF}, SiamRPN \cite{SiamRPN} and TRACA \cite{TRACA}. 
In Figure~\ref{fig:6} (right), we can observe that our methods achieve comparable or even better results compared with the recent low-efficiency deep trackers. It should be noted that most remarkable non-realtime trackers (e.g., VITAL \cite{VITAL} and MDNet \cite{MDNet}) cannot operate at a real-time speed even with the modern GPU device, but ours are CPU real-time.

\begin{figure}
	\centering
	\includegraphics[width=4.3cm]{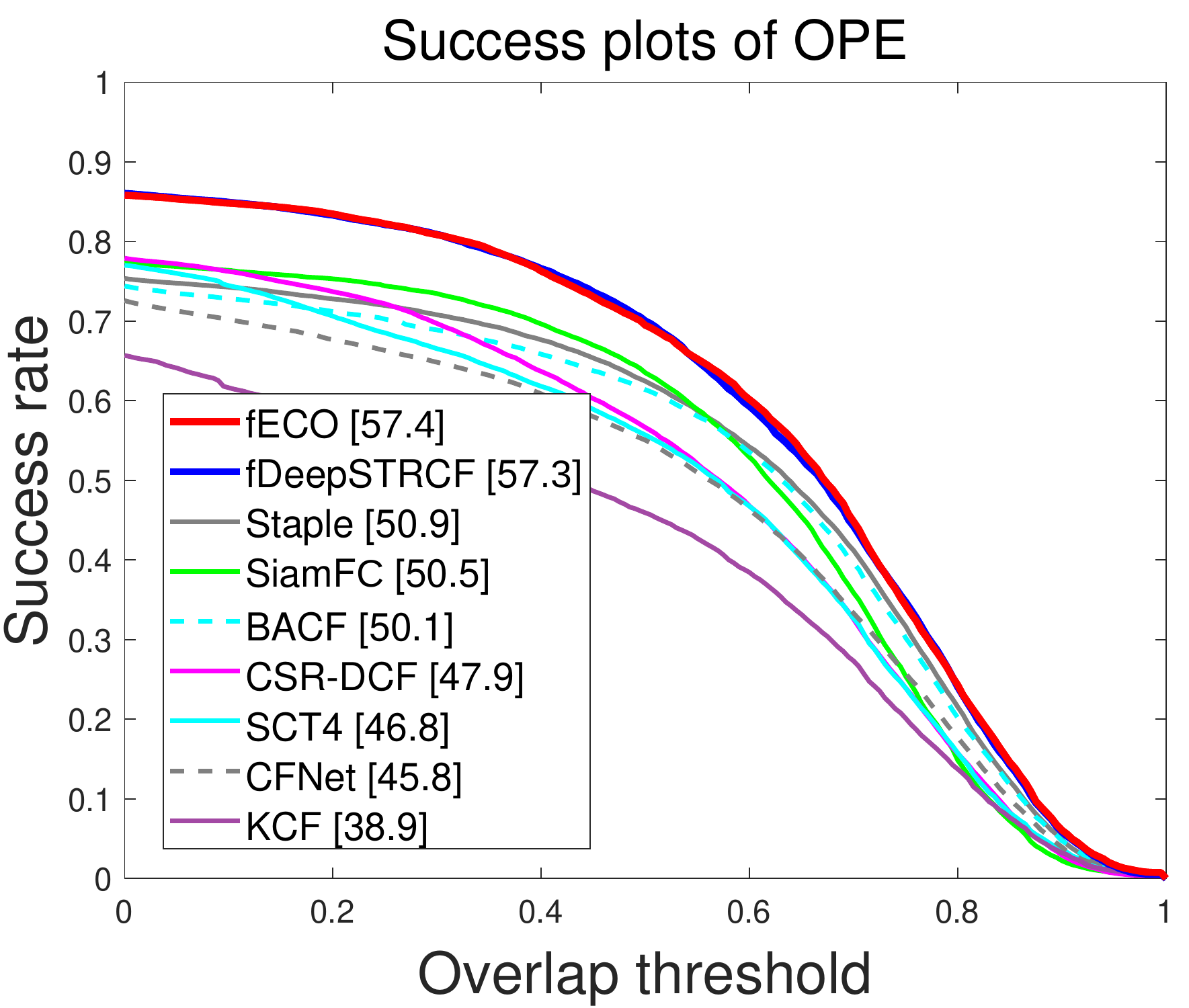}
	\includegraphics[width=4.3cm]{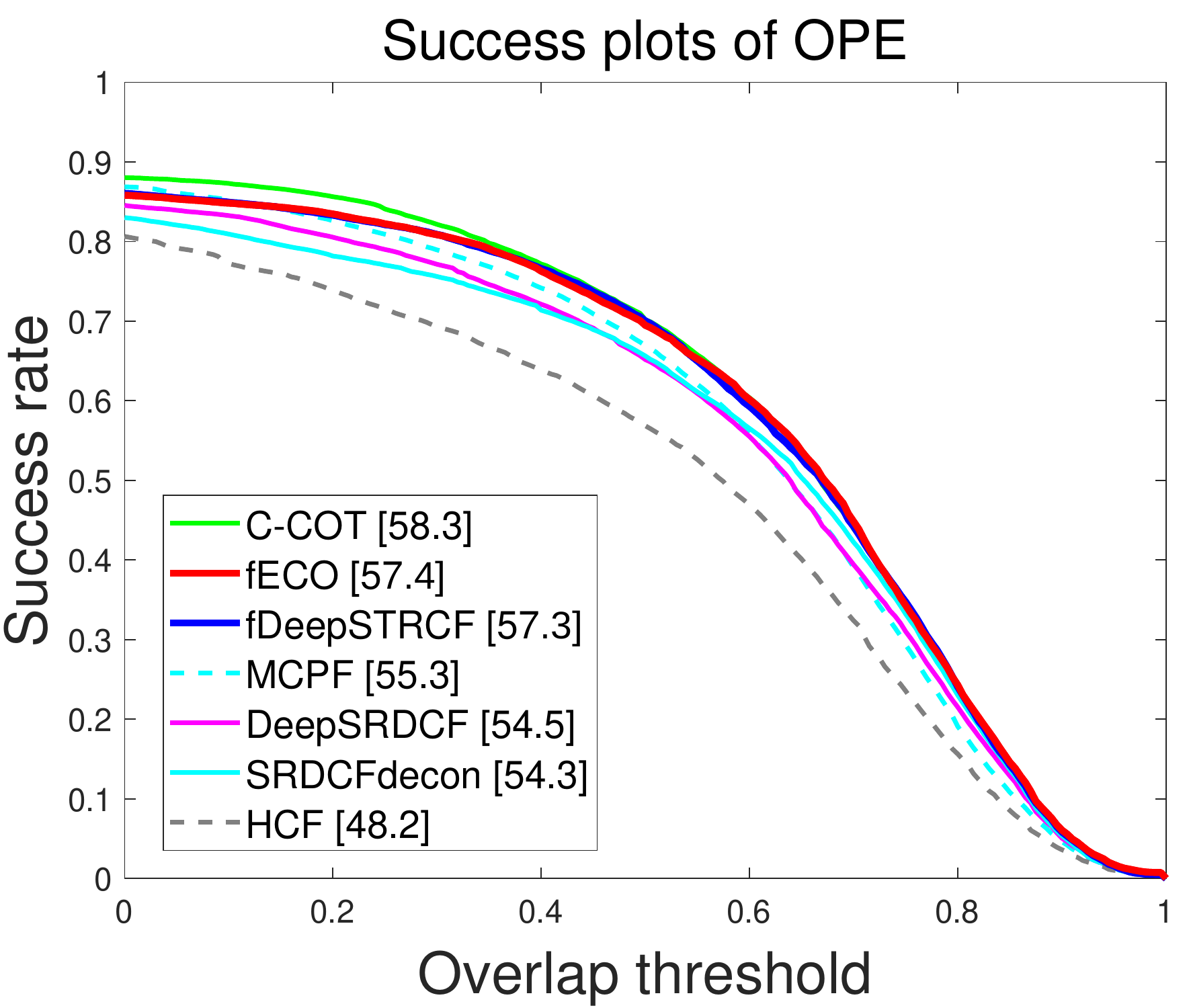}
	\caption{Success plots of real-time trackers (left) and non-realtime trackers (right) on the Temple-Color \cite{TempleColor128} dataset. Our trackers show outstanding performance among real-time trackers and comparably favorable results among non-realtime trackers.} \label{fig:8} \vspace{-0.0in}
\end{figure}

\setlength{\tabcolsep}{2pt}
\begin{table}
	\scriptsize
	\begin{center}
		\caption{The expected average overlap (EAO) of state-of-the-art methods on the VOT-2016 \cite{VOT2016} (left) and VOT-2017 \cite{VOT2017} (right) datasets. The comparative methods include the top performers on both datasets, our baseline methods (ECO \cite{ECO} and DeepSTRCF \cite{STRCF}) and the recently proposed trackers.} \label{table:table5}	
		\vspace{+0.0in}
		\begin{tabular}{cc}
			\begin{minipage}{0.25\textwidth}
				\begin{tabular*}{3.9 cm} {@{\extracolsep{\fill}}llc}
					\hline
					&Trackers & EAO   \\	
					\hline
					\multirow{8}{*}{\rotatebox{90}{Non-realtime}}
					&ECO \cite{ECO}   &0.374 \\
					&DSLT \cite{DSLT}  &0.332 \\
					&VITAL \cite{VITAL}  &0.323 \\
					&FlowTrack \cite{FlowCF}  &0.334 \\
					&DeepSTRCF \cite{STRCF}  &0.313 \\
					&C-COT \cite{C-COT}  &0.331 \\
					&MDNet \cite{MDNet}  &0.227 \\
					\hline
					\multirow{9}{*}{\rotatebox{90}{Real-time Trackers}}
					&SiamRPN \cite{SiamRPN}  &0.344 \\
					&SA-Siam \cite{SASiam}  &0.291 \\
					&StructSiam \cite{StructSiam}  &0.264 \\
					&MemTrack \cite{MemTrack}  &0.273 \\
					&ECOhc \cite{ECO}  &0.238\\
					&STRCF \cite{STRCF}  &0.279 \\
					&BACF \cite{BACF}  &0.233 \\
					&Staple \cite{Staple}  &0.295 \\
					&SiamFC \cite{SiamFc}  &0.277 \\
					\hline
					&{\noindent \bf fDeepSTRCF}  &0.308  \\
					&{\noindent \bf fECO}  &0.339  \\
					\hline
				\end{tabular*}
			\end{minipage}
			\hfil
			\begin{minipage}{0.5\textwidth}
				\begin{tabular*}{3.9 cm} {@{\extracolsep{\fill}}llc}
					\hline
					&Trackers & EAO   \\	
					\hline
					\multirow{8}{*}{\rotatebox{90}{Non-realtime}}
					&LSART \cite{LSART}  &0.323 \\
					&CFCF \cite{CFCF} &0.286\\
					&ECO \cite{ECO}   &0.280 \\
					&C-COT \cite{C-COT}  &0.267 \\
					&MCPF \cite{MCPF}  &0.248 \\
					&DeepSTRCF \cite{STRCF}  &0.227 \\
					&DLST \cite{VOT2017} &0.233\\
					\hline
					\multirow{9}{*}{\rotatebox{90}{Real-time Trackers}}
					&SiamRPN \cite{SiamRPN}  &0.243 \\
					&SiamDCF \cite{VOT2017} &0.249 \\
					&SA-Siam \cite{SASiam}  &0.236 \\
					&CSRDCF++ \cite{CSR-DCF}  &0.229\\
					&ECOhc \cite{ECO}  &0.238\\
					&STRCF \cite{STRCF}  &0.162 \\
					&UCT \cite{VOT2017} &0.206 \\
					&Staple \cite{Staple}  &0.169 \\
					&SiamFC \cite{SiamFc}  &0.188 \\
					\hline
					&{\noindent \bf fDeepSTRCF} &0.214 \\
					&{\noindent \bf fECO}  &0.255  \\
					\hline
				\end{tabular*}
			\end{minipage}
		\end{tabular}
	\end{center}
\end{table}
\vspace{-0.0in}

\begin{table*}
	\scriptsize
	\begin{center}
		\caption{Attribute-based evaluation on the OTB-2015 benchmark \cite{OTB-2015}. The evaluation metric is the area-under-curve (AUC) score of the success plot. The first and second highest values are highlighted by bold and underline.} \label{table:attribute1}
		\vspace{+0.0in}	
		\begin{tabular*}{15.5 cm} {@{\extracolsep{\fill}}lcccccccccccc}
			\hline
			& IV  & SV & OCC &DEF &MB &FM &IPR &OPR &OV &BC &LR &Overall \\
			\hline  
			~~TRACA \cite{TRACA}   &61.8 &56.8 & 57.1  &56.0 &58.7 &57.4 &58.0 &59.3 &56.5 &60.6 &50.5 &60.2\\
			~~SiamRPN \cite{SiamRPN}   &65.7 &62.0 &59.4  &60.8  &62.6 &59.8 &62.3 &62.3 &55.8 &60.9 &{\bf 67.8} &63.7\\
			~~BACF \cite{BACF}    &65.3 &58.8 & 58.4 &59.2 &59.5 &61.5 &59.1 &59.3 &56.0 &63.5 &52.0 &63.3\\
			~~CFNet \cite{CFNet}     &54.2 &54.7 & 51.6 &47.7 &54.5 &55.0 &56.9 &54.4 &41.9 &55.6 &63.5 &56.8\\
			~~CSR-DCF \cite{CSR-DCF}     &54.0 &52.0 & 53.8 &53.4 &58.4 &57.5 &51.1 &51.1 &51.0 &52.7 &44.4 &58.5\\
			~~ACFN  \cite{ACFN}    &56.5 &56.3 & 54.5 &53.6 &56.4 &57.0 &54.5 &54.6 &51.4 &54.8 &52.0 &57.0 \\
			~~SiamFC \cite{SiamFc}     &56.7 &56.6 & 54.3 &50.9 &54.6 &56.6 &55.7 &55.9 &51.4 &53.0 &63.0 &58.2\\
			~~Staple  \cite{Staple}    &60.4 &54.3 & 55.3 &55.9 &55.7 &55.1 &56.2 &54.8 &51.2 &59.0 &40.3 &59.1\\
			~~SCT4  \cite{SCT}    &52.6 &44.2 & 50.5 &51.2 &53.0 &54.1 &52.8 &51.8 &43.7 &55.6 &29.0 &53.8\\
			~~KCF  \cite{KCF}    &48.7 &39.7 & 44.9 &44.5 &46.8 &46.6 &47.6 &45.9 &39.7 &50.4 &28.8 &48.3\\
			\hline
			~~fECO      &{\bf 69.1} &\underline{65.4} &{\bf 66.2} &\underline{64.8} &\underline{67.3} &\underline{64.5} &\underline{62.8} &\underline{65.6} &{\bf 63.0} &{\bf 68.5} &54.1 &\underline{68.2}\\
			~~fDeepSTRCF  &\underline{68.5} &{\bf 66.9} &\underline{65.6} &{\bf 64.9} &{\bf 67.9} &{\bf 66.1} &{\bf 63.5} &{\bf 66.6} &\underline{62.4} &\underline{67.5} &\underline{64.5} &{\bf 68.6}\\
			\hline
		\end{tabular*}
	\end{center}
	\vspace{+0.0in}
\end{table*}

\begin{table*}
	\scriptsize
	\begin{center}
		\caption{Attribute-based evaluation between our methods and their corresponding baselines (ECO \cite{ECO} and DeepSTRCF \cite{STRCF}) with uncompressed networks. The AUC score is reported on the OTB-2015 dataset \cite{OTB-2015}.} \label{table:attribute2}
		\vspace{+0.0in}	
		\begin{tabular*}{15.5 cm} {@{\extracolsep{\fill}}lcccccccccccc}
			\hline
			& IV  & SV & OCC &DEF &MB &FM &IPR &OPR &OV &BC &LR &Overall \\
			\hline  
			~~ECO \cite{ECO}   &71.2 & 68.2 & 68.0  &63.4 &70.4 &68.1 &65.4 &67.5 &67.1 &71.2 &58.1 &69.4\\
			~~fECO      &69.1 &65.4 &66.2 &64.8 &67.3 &64.5 &62.8 &65.6 &63.0 &68.5 &54.1 &68.2\\
			~~~~$ \Delta $   &-2.1  &-2.8 &-1.8 &+1.4 &-3.1 &-3.6 &-2.6 &-1.9 &-4.1 &-2.7 &-4.0 &-1.2\\
			\hline
			~~DeepSTRCF \cite{STRCF}    &67.5 &66.8 &66.2 &64.1 &68.3 &66.5 &62.9 &66.6 &64.8 &64.6 &63.7 &68.5\\
			~~fDeepSTRCF  &68.5 &66.9 &65.6 &64.9 &67.9 &66.1 &63.5 &66.6 &62.4 &67.5 &64.5 &68.6\\
			~~~~$ \Delta $   &+1.0 &+0.1 &-0.6 &+0.8 &-0.4 &-0.4 &+0.6 &0 &-2.4 &+2.9 &+0.8 &+0.1\\
			\hline
		\end{tabular*}
	\end{center}
	\vspace{-0.0in}
\end{table*}

{\flushleft \bf OTB-2015 Dataset.} OTB-2015 is a popular tracking benchmark which extends the OTB-2013 dataset  with additional 50 challenging videos. On this dataset, our fECO and fDeepSTRCF exhibit the AUC scores of 68.2\% and 68.6\%, respectively. Figure~\ref{fig:7} shows that our methods outperform the recent real-time trackers and perform favorably against non-realtime deep trackers. The TRACA tracker \cite{TRACA} uses an encoder network to reduce the feature channel and achieves high speed on GPU. In contrast, our CF-VGG not only reduces feature dimension but also greatly accelerates the feature extraction time, which brings in real-time speed on the CPU and better performance (about 8\% higher in AUC). The recent SiamRPN \cite{SiamRPN} improves the SiamFC tracker \cite{SiamFc} and achieves impressive performance. However, it needs GPU to achieve high speed and our CPU real-time methods still outperform it by about 5\% in  AUC score. The deep feature representation of CF-VGG enables our trackers to surpass traditional CF trackers using empirical features (e.g., BACF \cite{BACF}, Staple \cite{Staple}, and CSR-DCF \cite{CSR-DCF}). Furthermore, our methods even outperform many recent deep trackers that run at only 1 FPS on GPU (e.g., VITAL \cite{VITAL} and MDNet \cite{MDNet}).

{\flushleft \bf Temple-Color Dataset.} We further evaluate our trackers on the Temple-Color benchmark with 128 color videos. On the Temple-Color, our fECO and fDeepSTRCF yield the AUC scores of 57.4\% and 57.3\%, respectively. From the left figure in Figure~\ref{fig:8}, we can observe that our trackers perform better than  state-of-the-art real-time trackers (e.g., BACF \cite{BACF}, Staple \cite{Staple} and SiamFC \cite{SiamFc}). Compared with the non-realtime deep trackers including C-COT \cite{C-COT} and MCPF \cite{MCPF}, the improved fECO and fDeepSTRCF trackers achieve comparable performance.

{\flushleft \bf VOT-2016 and VOT-2017 Datasets.} Finally, we compare our trackers with state-of-the-art methods on the VOT-2016 \cite{VOT2016} and VOT-2017 \cite{VOT2017} benchmarks. On the VOT benchmark, a tracker will be re-initialized when tracking failure occurs. The expected average overlap (EAO) is the evaluation metric which considers both the tracking accuracy (overlap with the ground truth box) and robustness (failure times) \cite{VOTpami}. As shown in Table \ref{table:table5}, our methods obviously outperform ECOhc and STRCF. This affirms that CF-VGG performs favorably against empirical features. In addition, our trackers achieve comparable or even better results than the VOT-2016 top performer C-COT \cite{C-COT}, whose running speed is only 0.3 FPS on a CPU.
Compared with other state-of-the-art and recently proposed trackers (e.g., SA-Siam \cite{SASiam}, VITAL \cite{VITAL}, DSLT \cite{DSLT}, SiamRPN \cite{SiamRPN}, FlowTrack \cite{FlowCF}), our methods overall show competitive performance.

\begin{figure}
	\centering
	\includegraphics[width=9cm]{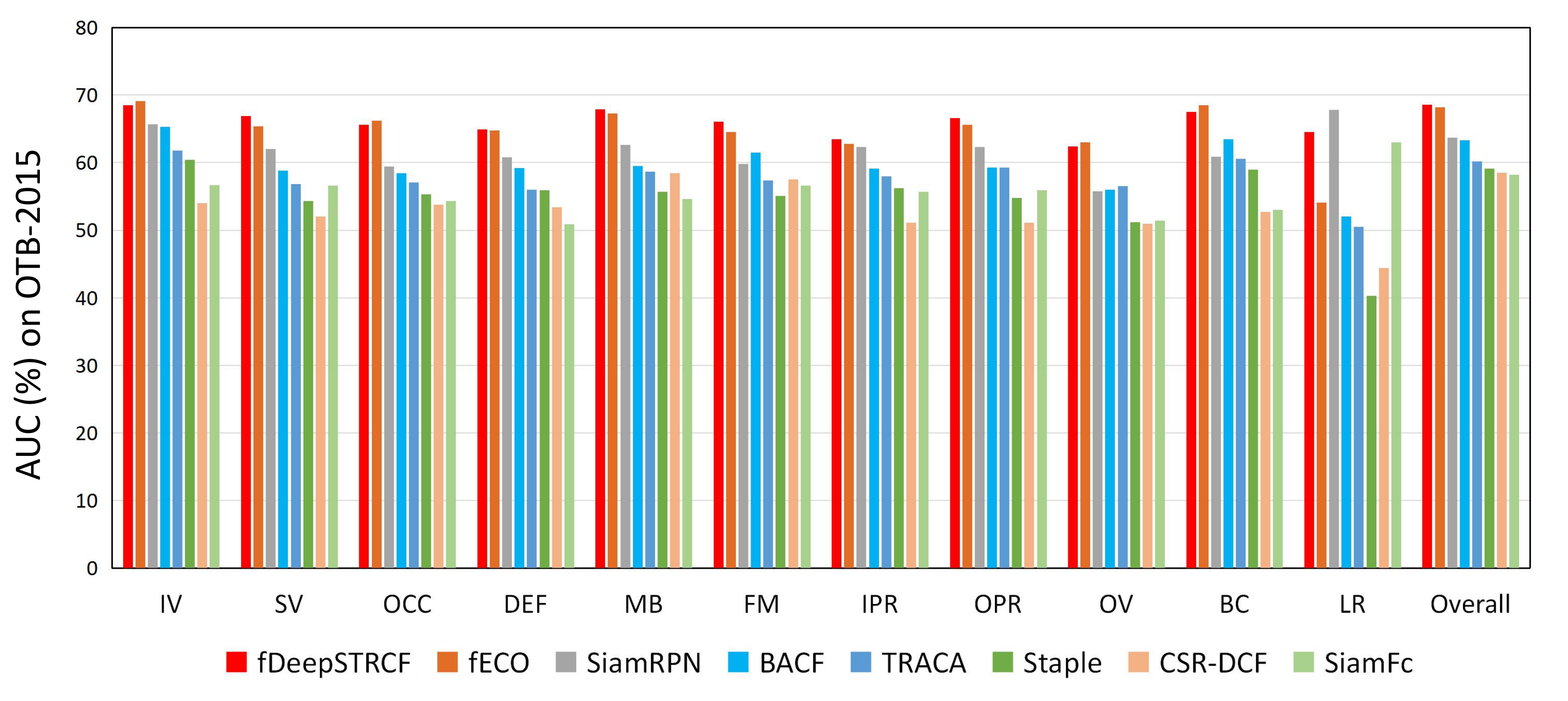}
	\caption{Attribute-based evaluation on the OTB-2015 benchmark \cite{OTB-2015}. The evaluation metric is the area-under-curve (AUC) score of the success plot. We also put the overall performance here (the last one) for comparison convenience facing a single challenge and their combination. Only the top 6 real-time trackers are displayed for clarity. Our fDeepSTRCF and fECO algorithms perform favorably against state-of-the-art real-time trackers in various challenging scenes.}
	\label{fig:attribute_fig} \vspace{+0.0in}
\end{figure}

{\flushleft \bf Attribute Evaluation.} All the 100 videos in OTB-2015 \cite{OTB-2015} are annotated with 11 different attributes, namely: background clutter (BC), deformation (DEF), out-of-plane rotation (OPR), scale variation (SV), occlusion (OCC), illumination variation (IV), motion blur (MB), in-plane rotation (IPR), out of view (OV), fast motion (FM) and low resolution (LR).

\begin{figure*}
	\centering
	\includegraphics[width=18.5cm]{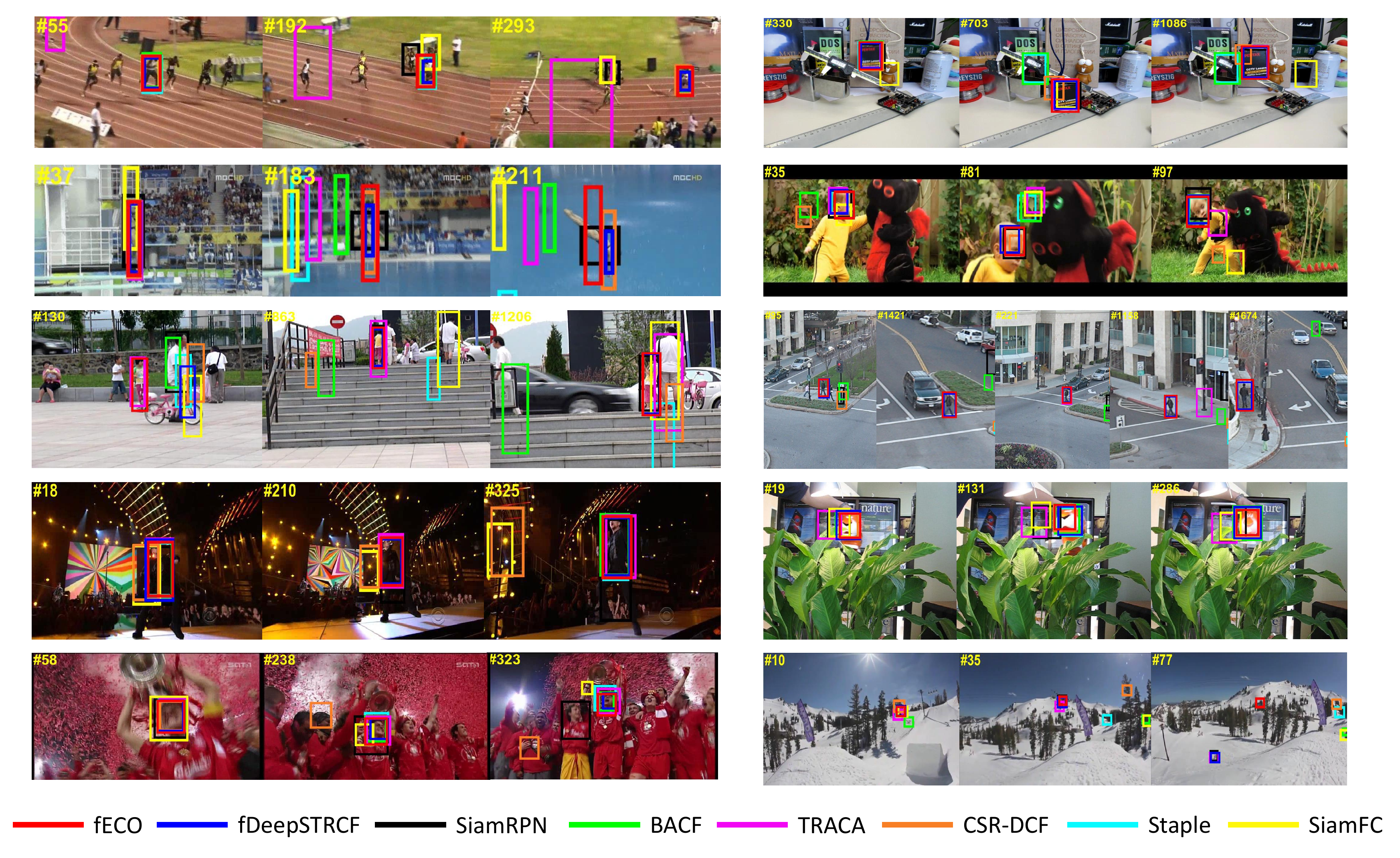}
	\vspace{-0.1in}
	\caption{Qualitative evaluation of our trackers (e.g., fECO, fDeepSTRCF) and six other state-of-the-art real-time trackers including SiamRPN \cite{SiamRPN}, BACF \cite{BACF}, TRACA \cite{TRACA}, CSR-DCF \cite{CSR-DCF}, Staple \cite{Staple} and SiamFC \cite{SiamFc} on 10 challenging sequences (from left to right and top to down: \emph{Bolt2, Box, Diving, DragonBaby, Girl2, Human3, Singer2, Tiger1, Soccer} and \emph{Skiing}, respectively). Our fECO and fDeepSTRCF trackers perform favorably against the state-of-the-arts.}
	\label{fig:qualitative} \vspace{+0.0in}
\end{figure*}

\begin{figure*}[htb]
	\centering
	\includegraphics[width=17.5cm]{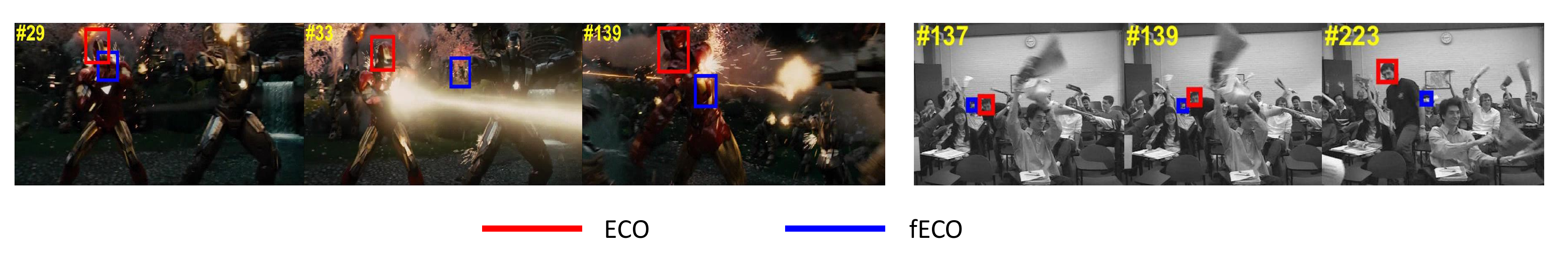}
	\vspace{-0.1in}
	\caption{Failure cases of the proposed method. The videos are \emph{Ironman} and \emph{Freeman4} from OTB-2015 \cite{OTB-2015}. Our compressed model struggles when an occlusion or a drastic appearance change occurs.}
	\label{fig:failure_case} \vspace{-0.0in}
\end{figure*}

In Table \ref{table:attribute1} and Figure \ref{fig:attribute_fig}, we show the comparison results of 10 real-time trackers (these trackers are from Section \ref{state-of-the-art comparison}) when facing the above challenging factors. The results show that our fECO and fDeepSTRCF trackers obviously outperform other competitors in almost all the challenging scenes.

In Table \ref{table:attribute2}, we further compare our methods with their teachers (i.e., ECO and DeepSTRCF) on attributed videos. From the results, we can observe that our compressed model is comparable with the teacher model in most attributes, but performs not good enough in fast motion (FM), motion blur (MB), out of view (OV) and low resolution (LR), which indicates the representation capability of our student network still has improvement room. It should be noted that the model size of our network is only 1/63 of its teacher, so the slight performance degradation is bearable since our trackers achieve superiorly balanced high performance and CPU real-time efficiency.

{\flushleft \bf Qualitative Evaluation.} Figure \ref{fig:qualitative} shows some comparison results of our trackers (fECO and fDeepSTRCF) and other six state-of-the-art real-time trackers including SiamRPN \cite{SiamRPN}, BACF \cite{BACF}, TRACA \cite{TRACA}, CSR-DCF \cite{CSR-DCF}, Staple \cite{Staple} and SiamFC \cite{SiamFc} on ten challenging sequences. From the results in Figure \ref{fig:3}, we can see that our fECO and fDeepSTRCF trackers perform well on occlusion (e.g., \emph{Box, Girl2, Human3} and \emph{Soccer}) and background clutter (e.g., \emph{Tiger1} and \emph{Soccer}). Compared with the recent real-time deep trackers (SiamRPN \cite{SiamRPN} and TRACA \cite{TRACA}), our methods perform favorably against them while exhibiting the CPU real-time speed.

\subsection{Failure Cases}

Finally, we show some failure cases of our method in Figure \ref{fig:failure_case}. In the video \emph{Freeman4}, the target with low resolution undergoes frequent occlusions in a short span of time, while the \emph{Ironman} in the second video occurs a drastic appearance change. In these cases, our compressed model is not powerful enough compared to the teacher network. In our future work, we aim to include more training data and adopt a better network structure to further enhance the representation capability of the student model.

\section{Conclusion}\label{sec:conclusion}

In this paper, we propose to learn a lightweight backbone network for real-time correlation tracking. By simultaneously compressing and transferring the teacher network pretrained on object recognition, we obtain a highly compressed lightweight model (63$\times$ smaller) as the feature backbone. Extensive experiments demonstrate that our training scheme and strategies are effective and efficient. Even though being extremely lightweight, the proposed distilled backbone network is sufficiently powerful and almost maintains the same feature representation capability as the teacher network. Leveraging our lightweight model for deep correlation tracking, the recent top CF trackers consume much less memory storage and show superiorly balanced high performance and CPU real-time efficiency.

\bibliographystyle{IEEEtran}
\bibliography{reference_full}









\end{document}